\documentclass[ twoside]{hstemplate}
\usepackage{blindtext}
\usepackage{amssymb}
\usepackage{amsmath}
\usepackage{graphicx}
\usepackage{soul}
\usepackage{xcolor}
\usepackage{algpseudocode}
\usepackage{algorithm}
\usepackage{booktabs}

\newcommand{\KL}[2]{\text{KL}\left( #1 \left| \vphantom{#2} \right| #2 \right)}

\newcommand{\expectation}[2]{\mathbb{E}_{#1}\AdaRectBracket{#2}}
\newcommand{\AdaBracket}[1]{\left(#1\right)}
\newcommand{\AdaRectBracket}[1]{\left[#1\right]}

\newcommand{\entropyany}[1]{\mathcal{H}\left( #1 \right)}

\usepackage[most]{tcolorbox}

\newtcolorbox[use counter=conclusion]{ourconclusion}{%
	blanker, borderline west={3pt}{0pt}{blue!40!black},
	left=10pt, top=4pt, bottom=4pt,
	before skip=8pt, after skip=8pt,
	fontupper=\itshape,
	before upper={\textbf{\textcolor{blue!40!black}{Conclusion~\thetcbcounter.}}\space}
}

\newif\ifcomments

\commentsfalse   

\ifcomments
\newcommand{\haseeb}[1]{\textcolor{purple}{haseeb: #1}}
\newcommand{\martha}[1]{\textcolor{teal}{martha: #1}}
\newcommand{\lingwei}[1]{\textcolor{brown}{lingwei: #1}}
\newcommand{\adam}[1]{\textcolor{red}{adam: #1}}
\else
\newcommand{\haseeb}[1]{}
\newcommand{\martha}[1]{}
\newcommand{\lingwei}[1]{}
\newcommand{\adam}[1]{}
\fi

\usepackage{enumitem}

\newlist{conclusions}{enumerate}{1}
\setlist[conclusions]{
	font=\sffamily\small\bfseries,
	label={C\arabic*:},
	leftmargin=*,
	itemsep=0.6\baselineskip,
	topsep=0.4\baselineskip
}

\newlist{concldesc}{description}{1}
\setlist[concldesc]{
	leftmargin=*,
	itemsep=0.6\baselineskip,
	topsep=0.4\baselineskip,
	font=\normalfont, 
	style=nextline
}

\newcommand{\pparams}{\theta}
\newcommand{\qparams}{w}

\newcommand{\trueq}{r}

\usepackage{tocloft}

\leadauthor{H. Shah} 

\begin{document}

\title{Deconstructing Actor-Critic: A Large-scale Empirical Study of Design Components for Practitioners}
\shorttitle{Deconstructing Actor-Critic}
\author[a,b]{Haseeb Shah\thanks{To whom correspondence should be addressed. E-mail: hshah1@ualberta.ca}\thanks{H.S. contributed equally to this work with L.Z.}}
\author[c]{Lingwei Zhu\samethanks}

\author[a,b,d]{Adam White}
\author[a,b,d]{Martha White}

\affil[a]{Department of Computing Science, University of Alberta, Canada}
\affil[b]{Alberta Machine Intelligence Institute (amii), Canada}
\affil[c]{Great Bay University, China}
\affil[d]{Canada CIFAR AI Chair}

\maketitle

\begin{abstract}
Reinforcement learning is increasingly being considered for controlling real-world systems, from fusion plasma and autonomous vehicles to drug discovery and drinking water treatment, where reliability is essential and tuning budgets are limited.
Actor-critic algorithms share a set of design decisions, such as how the policy is updated, how it represents the distribution over actions, how its gradient is estimated, and how often it is updated relative to the value estimator.
Using a control task derived from a real water treatment plant, we analyze over 33,000 experiments to determine how these components affect variability across runs and sensitivity to hyperparameters.
Common defaults, such as Gaussian action distributions with pathwise gradient estimators, are among the least reliable configurations, whereas bounded distributions with adaptive update schedules remain robust across a wide range of settings.
These findings offer empirical guidance to practitioners across scientific and engineering domains for understanding and making component-level decisions when adapting actor-critic methods to new real-world control settings.
	\end{abstract}
\begin{keywords}
\textbf{Keywords:} reinforcement learning; actor-critic algorithms; process control; algorithm reliability
\end{keywords}
\vspace{1em}
	\dropcap{R}einforcement Learning (RL) algorithms are increasingly being used in domains ranging from racing games and optimizing computer operations to simulated reactor control \citep{Silver2017-GowithoutHumanKnowledge,Fawzi2022-matrixDRL-nature,Wurman2022-DRL-driving-nature,Deepmind-2022-RLtokamak,Feng-2023-nature-RLAutoDrive,wang2025-surveyRLsoftware}, but real-world deployments are much more rare because current popular algorithms are designed for and tuned to specific academic benchmarks. Deep RL algorithms are actually a collection of specific components that do not generally work well across problems. For example, the Rainbow agent~\citep{hessel2018rainbow} combines several algorithmic ideas (noisy-net exploration~\citep{fortunato2018noisy}, distributional losses~\citep{bellemare2017distributional}, prioritized experience replay~\citep{schaul2015prioritized}, etc.) to achieve near state-of-the-art on Atari~\citep{bellemare2013arcade}, but performs poorly in small-scale classic control environments like Mountain Car~\citep{ceron2021revisiting}. Soft Actor-Critic combines reparameterized updates, squashed Gaussian policies, and entropy regularization~\citep{haarnoja-SAC2018} to great effect in MuJoCo tasks~\citep{todorov2012mujoco}, but generally performs poorly in the DM Control Suite~\citep{tassa2018deepmind}---both popular benchmarks for continuous-action control methods in RL. This lack of generality is a major obstacle to applying these algorithms to new problems and real-world deployments.     
	
	Actor-critic methods are of particular interest, as they can be applied to both discrete- and continuous-action problems \citep{abdolmaleki2018maximum,haarnoja-SAC2018,Neumann2023-greedyAC,Zhu2023-tsallisOffline} and form the backbone of many of the most widely used RL systems.
	At a high level, these methods share a common template: a policy responsible for selecting actions (actor) and a learned estimator that evaluates actions (critic), enabling policy updates with lower variance and greater scalability than vanilla policy gradient algorithms.
	Despite this common backbone, these algorithms differ in how they define and implement policy improvement, for example, through conservative, constraint-based updates \citep{trpo-schulman15,schulman2017proximal}, regularized objectives that encourage exploration \citep{haarnoja-SAC2018}, or improvement and projection procedures \citep{abdolmaleki2018maximum}. 
	These differences can interact strongly with how the policy is parameterized and optimization choices, resulting in substantially different qualitative behaviors even though all these methods are grouped under the same umbrella of ``actor-critic" algorithms.
	
Our limited understanding of individual algorithmic components and the different ways to implement them limits our ability to deploy them effectively.
    There is a dizzying array of algorithms to choose from, with insufficient clarity on their strengths, weaknesses, and failure modes.
	The behavior of an algorithm is often inseparable from the bundle of implementation tricks and tuning decisions such as normalization choices, entropy schedules, clipping and regularization strengths, actor and critic update ratios, target networks, replay buffer sizes and sampling strategies, and learning rate schedules.
	These choices can significantly change performance and even reverse conclusions across empirical studies.
	As a result, a common strategy is to adopt complete implementations of popular algorithms, such as PPO \citep{schulman2017proximal} or DDPG \citep{Lillicrap2016-ddpg}, and avoid modifying them, even when the target problem differs from the setting for which the implementation was originally designed. 
	This can lead to awkward workarounds that preserve code structure rather than adapting the underlying algorithm to the problem at hand. 
    For example, to apply DDPG to bandit problems, researchers have had to rely on awkward state padding to preserve the original code structure \citep{eslam2025simulate}.

	\begin{figure*}[ht!]
		\centering
		\includegraphics[width=\linewidth]{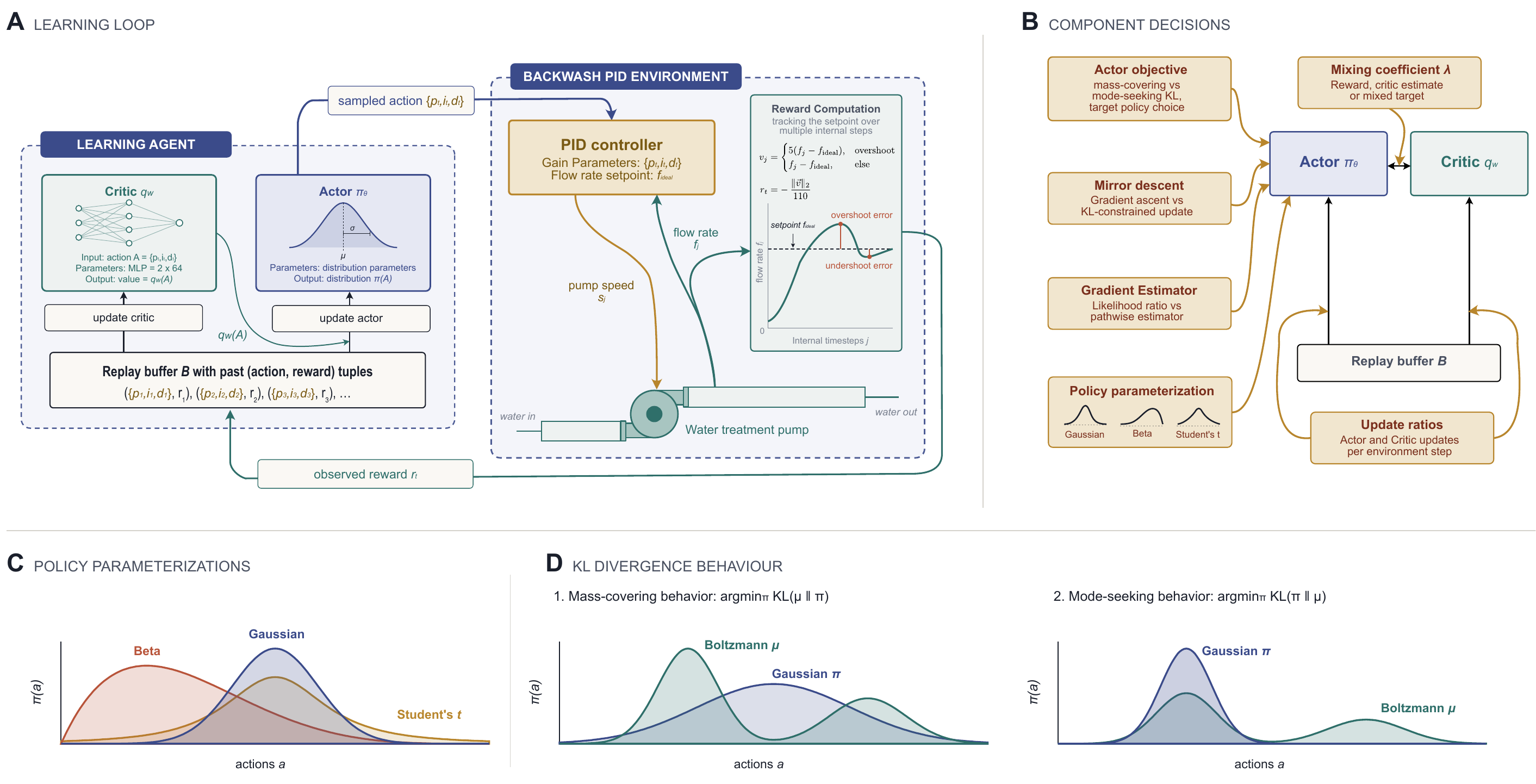}
		\caption{
			\textbf{Schematic of a reinforcement learning system on a water treatment controller}.
			\textbf{A}: Deployment setting and agent internals.
			The actor is a probability distribution over PID gain parameters ($p_t, i_t, d_t$), parameterized here by the mean $\mu$ and standard deviation $\sigma$.
			An RL agent containing this actor $\pi_\theta$, a critic $q_w$ and a replay buffer $B$ of past data, outputs PID gains and writes them to a classical PID controller.
			This PID controller drives the backwashing pump in a water treatment system by controlling the pump speed $s_j$ over $j \in \{1, \dots, \mathcal J\}$ discrete internal timesteps for $\mathcal J = 55$, where each step represents 1 second of water flow through the pump. The goal is to achieve a water flow rate $f_j$ equal to the setpoint $f_{\text{ideal}}$.
			The tracking error $f_j -f_{\text{ideal}}$ is fed back into the PID controller, forming a closed feedback loop.
			Finally, the reward $r_{t}$ is constructed by summarizing the tracking quality over $\mathcal J$ steps and the pair $(a_t = (p_t, i_t, d_t), r_t)$ stored in the agent's buffer.
			The actor and critic sample from this buffer to do their updates, described in more detail below. 
			\textbf{B}: We run over 33,000 experiments and analyze variations within six different components of the learning agent to study the reliability of the system.
			\textbf{C}: A visualization of some of the policy parameterizations available to the actor. The Student's $t$ policy can achieve heavier tails compared to the Gaussian, while the beta policy is naturally bounded.
			\textbf{D}: We later investigate the impact of using the mass-covering KL vs the mode-seeking KL, and so visualize the difference between the two here. The visualization shows the outcome from optimizing a Gaussian policy $\pi$ to approximate a bimodal Boltzmann target $\mu$.
			1. Minimizing the mass-covering KL results in a learned $\pi$ that is spread over both modes.
			2. Minimizing mode-seeking KL instead commits the learned $\pi$ to a single mode. 
		}
		\label{fig:hero}
	\end{figure*}
    
	The goal of this paper is to provide clarity on the key design components and implementation choices in actor-critic algorithms, both conceptually and through an empirical study.
	Such an analysis can make it easier for both practitioners and researchers to select, modify, and build on actor-critic algorithms, particularly those who care about achieving reliable performance.
	Rather than focusing solely on best-case performance, we emphasize both the variability of performance across runs and hyperparameter sensitivity, since these factors determine whether an algorithm is usable in practice and can be adapted to a new domain with a reasonable tuning budget and minimal modifications.
	
	To isolate the effects of actor-critic algorithms, we study a simplified, stateless PID tuning setting. 
	Tuning PID controllers is a popular task in engineered systems where small changes in parameters can have large effects on overall stability and efficiency.
	This setting strips away many of the confounding factors that dominate outcomes in deep RL benchmarks, such as long-horizon credit assignment, environment-specific engineering, and temporal state dependence, to name a few. This simplified setting allows us to explain the key differences between these algorithms while also making it easier to run comprehensive hyperparameter sweeps that both provide insights for existing algorithms and highlight promising avenues for future study. Our goal is to provide empirical insights that are transferable 
    through understanding rather than performance-based benchmarking.  
	
	In particular, we investigate three key choices: how the actor (policy) is updated, how the actor is parameterized, and the utility of typical optimization improvements to the update (e.g., mirror descent). We focus on several widely-used actor-critic algorithms: Proximal Policy Optimization (PPO) \citep{schulman2017proximal}, Soft Actor-Critic (SAC) \citep{haarnoja-SAC2018}, Deep Deterministic Policy Gradient (DDPG) \citep{Lillicrap2016-ddpg}, Maximum a Posteriori Policy Optimization (MPO) \citep{abdolmaleki2018maximum}, Greedy Actor-Critic (GreedyAC) \citep{Neumann2023-greedyAC} and two simple methods, vanilla Actor-Critic and REINFORCE \citep{Williams-reinforce}. Throughout, we focus on hyperparameter sensitivity, which is critical for ease of use and for practitioners to feasibly use and deploy these algorithms. We introduce a PID tuning environment based on real data from a drinking water treatment pump controller. 
	Even in this simple bandit setting, we observe differences among methods and identify trends linking stability to policy parameterizations and to how updates are performed.

	\section*{Problem setting and base algorithms}
	We restrict our focus to a non-contextual continuous-action bandit setting, to make the key ideas behind these actor-critic algorithms more accessible and to facilitate a more thorough empirical analysis. 
	In the bandit setting, on each time step $t$, the agent takes action $A_t \in \mathcal{A}$ and then observes the outcome reward $R_t \in \mathbb{R}$.\footnote{In the full reinforcement learning setting, this reward is typically labeled $R_{t+1}$, because it might depend on the next state. We chose to use $R_t$ to be clear it is the reward from taking action $A_t$.} The outcome reward is given by $R_t = r(A_t) + \epsilon$ for reward function $r: \mathcal{A} \rightarrow \mathbb{R}$ and some independent noise $\epsilon$. The agent's goal is to maximize accumulated rewards, in other words, its average reward $\frac{1}{t} \sum_{i=1}^t R_i$, and ultimately find a policy $\pi$ with high expected reward, $\mathbb{E}_\pi[R]$. 
	
	To do so, the agent learns a parameterized policy $\pi_\theta : \mathcal{A} \rightarrow [0, \infty)$ with parameters $\theta$. In the continuous action setting, $\theta$ is the parameters for a probability density function (pdf), such as a Gaussian (with $\theta = (\mu, \sigma)$ for mean $\mu$ and variance $\sigma^2$), a beta distribution (with $\theta = (\alpha, \beta)$ for scale parameters $\alpha >0, \beta > 0$) and other options discussed in more detail in SI Appendix \ref{app:params}. The goal is to find the best parameterized policy in the policy class (given by $\Theta$) 
	%
	\begin{equation}
		\max_{\theta \in \Theta} \mathbb{E}_{\pi_\theta}[R] = \int_{\mathcal{A}} \pi_\theta(a) r(a) \mathrm{d}a \label{eq_obj}
	\end{equation}
	where parameter set $\Theta$ is bounded to prevent the policy from becoming fully deterministic. For example, for a Gaussian, we might have $\Theta = \{(\mu, \sigma) \in \mathbb{R}^2 \ | \ \sigma^2 > 0.001 \}$. The overall learning loop for the bandit task we consider in this paper is depicted in Figure \ref{fig:hero}(A). 
	
	A basic algorithm to achieve this goal is REINFORCE \citep{Williams-reinforce}, which uses a (stochastic) gradient ascent update based on the objective in Equation \ref{eq_obj}:
	\begin{align*}
		\nabla_\pparams \mathbb{E}_{\pi_\pparams}[\trueq(A)] &=\nabla_\pparams \int_{\mathcal{A}} \pi_\pparams(a) r(a) \mathrm{d}a
		= \int_{\mathcal{A}} r(a) \nabla_\pparams \pi_\pparams(a)  \mathrm{d}a\\
		&= \int_{\mathcal{A}} r(a) \pi_\pparams(a) \nabla_\pparams \ln \pi_\pparams(a) \mathrm{d}a \\
		&= \mathbb{E}_{\pi_\pparams}[\trueq(A)  \nabla_\pparams \ln \pi_\pparams(A) ] 
	\end{align*}
	This gradient increases the probability of high-reward actions and decreases it for low-reward actions.
	To accelerate learning, it is standard to subtract a baseline $v$ from $r(A)$, to center values around zero.
	It is equivalent to use $r(A) - v$ in the update because 
	%
	\begin{align*}
		\nabla_\pparams \mathbb{E}_{\pi_\pparams}[\trueq(A) - v] &=
		\nabla_\pparams \mathbb{E}_{\pi_\pparams}[\trueq(A)]  - \underbrace{\nabla_\pparams \mathbb{E}_{\pi_\pparams}[v]}_{=0}.
	\end{align*}
	That is, any baseline that does not depend on the action will have zero contribution to the gradient.
	The baseline is typically chosen to be $v \approx \mathbb{E}_{\pi_\pparams}[\trueq(A)]$.
	Therefore, we have the advantage of an action in the update $\trueq(A) - v$. This way, the gradient has a more intuitive meaning: the probabilities of actions with positive values will increase, and those with negative values will decrease.
	
	REINFORCE uses a stochastic sample of this update on time step $t$, taking action $A_t \sim \pi_\theta$, observing reward $R_t$, and updating with
	\begin{align*}
		\pparams &\gets \pparams + \eta (R_t - v) \nabla \ln \pi_\pparams(A_t) && \textbf{(REINFORCE)}\\
		v &\gets (1-\eta_v) v + \eta_v R_{t} &&
	\end{align*}
	for some stepsizes $\eta, \eta_v > 0$ and initializing $v = 0$ at $t = 0$.
	
	REINFORCE, however, is known to suffer from high-variance; instead we can consider \emph{actor-critic} (AC) methods that estimate $q_w(a) \approx r(a)$ instead of only using sampled rewards. As the agent generates interactions $(A_t, R_{t})$, it can store them in a replay buffer $\mathcal{B} = \{(A_1, R_1), (A_2, R_2), \ldots, (A_t, R_t)\}$ up to the current time $t$. It can use a regression algorithm to update $q_\qparams$ using the dataset $\mathcal{B}$, with the primary goal of estimating $\trueq$ as accurately as possible. For example, the parameterized $q_w$ can be a neural network, with parameters $w$, updated using mini-batch SGD. The update can use $q_w$ instead of a (noisy) sampled $R_t$
	\begin{align*}
		\pparams &\gets \pparams + \eta (q_w(A_t) - v) \nabla \ln \pi_\pparams(A_t) && \textbf{(AC)}
	\end{align*}
    This buffer-based approach forms the basis of the actor-critic algorithms we consider, as depicted in Figure \ref{fig:hero} (A). 
    
	However, this update can be \emph{biased}, since it is likely that $q_w(a) \neq \mathbb{E}[R_t | A_t = a]$. To mitigate this, a parameter $\lambda \in [0,1]$ can be introduced that mixes between the unbiased, but high variance $R_t$ and the low variance, but potentially biased $q_w(a)$ 
	\begin{align*}
		\!\pparams &\gets \pparams \!+\! \eta (\lambda R_t + (1\!-\!\lambda) q_w(A_t) - v) \nabla \ln \pi_\pparams(A_t)  \  \textbf{(AC($\lambda$))}
	\end{align*}
	When $\lambda = 1$, we recover REINFORCE.
	
	These two base algorithms, REINFORCE and AC(0), provide two canonical policy optimization approaches. REINFORCE is a \emph{policy gradient} method, using an (unbiased) stochastic gradient update on the policy objective. AC(0), on the other hand, can be interpreted as approximate policy iteration \citep{chan2021-greedification}. It alternates between approximate evaluation (updating $q_w$) and approximate greedification (updating $\pi_\theta$) to concentrate on high-value actions according to $q_w$. 
	In the bandit setting, approximate evaluation is simply a regression problem, and we opt to use the same strategy across methods (see Alg. \ref{alg:critic}). 
	However, the core ideas behind different actor updates (approximate greedification) persist in the bandit setting, as they greedify differently given $q_w(a)$. In this work, we focus on understanding the differences in these actor updates across variants of algorithms that build on AC(0), and also present experiments contrasting the use of algorithms like REINFORCE with AC(0). 
	
	\section*{Key Design Decisions in the Actor Update}
	
	In this section, we outline several key design decisions for the actor update, summarized in Figure \ref{fig:hero} (B). We ground this discussion by focusing on five algorithms that are modifications of the basic REINFORCE and AC(0) algorithms described above: Proximal Policy Optimization (PPO), Soft Actor-Critic (SAC), Deep Deterministic Policy Gradient (DDPG), Maximum a Posteriori Policy Optimization (MPO), and Greedy Actor-Critic (GreedyAC). These algorithms are chosen because they cover several key differences in the actor update. PPO can be seen as a modification of REINFORCE that uses only estimates of $v$ without learning $q_w$. The remaining four algorithms can be viewed as modifications of AC(0) that rely heavily on $q_w$. 
	We provide pseudocode and details for each algorithm in SI Appendix \ref{sec:loss} and here describe the high-level conceptual differences. 
	
	\subsection*{Using $\lambda > 0$ or $\lambda = 0$}
	This is a key distinction between algorithms that are typically called \emph{on-policy} and \emph{off-policy}. When we set $\lambda = 0$, we use only $q_w$ and can counterfactually (off-policy) reason about any action. This decouples the actor update from the sampled action $A_t$ and allows us to leverage buffers of stored data to improve sample efficiency. This partially explains why there are so many off-policy AC algorithms, including the ones we consider here: SAC, DDPG, MPO and GreedyAC. These algorithms rely on estimates of the value of each action, and their key differences lie in trying to develop an improved actor update (approximate greedification). 
	
	Once we pick $\lambda > 0$, we use $R_t$ that resulted from sampling $A_t$, giving an on-policy update. REINFORCE, AC($\lambda$) and PPO all use such updates. The  REINFORCE update above use only the most recent sample, whereas PPO performs multiple epochs over a recently sampled trajectory. The update becomes slightly off-policy after each update because the policy has changed, and PPO introduces an importance-sampling correction to account for this. However, these updates remain nearly on-policy, as the policy does not change significantly before this data is discarded and a new trajectory is generated under the current policy (due to clipping and other measures). 
	
	\subsection*{Using Mirror Descent and Trust-Regions}
	
	Another key choice in many actor-critic algorithms is to replace this standard gradient ascent update with a mirror ascent update. The gradient ascent update is obtained by using a local Taylor series approximation around the objective $J(\pparams) \doteq \mathbb{E}_{\pi_\pparams}[\trueq(A)]$ for the current point $\pparams_t$, with an Euclidean distance 
    \citep{beck2003mirror}
	\begin{align*}
		\hat{J}(\pparams) &\doteq J(\pparams_t) + (\pparams-\pparams_t)^\top \nabla J(\pparams_t) - \frac{1}{2 \eta}  \|\pparams-\pparams_t \|_2^2\\
		\pparams_{t+1} &= \arg\max_{\pparams} \hat{J}(\pparams) = \pparams_t + \eta \nabla J(\pparams_t)
	\end{align*}
	We get a closed-form update when maximizing $\hat{J}(\pparams)$ because $\hat{J}(\pparams)$ is quadratic. But an Euclidean distance is not always the right choice. For learning policies, it is more sensible to use the distance between distributions rather than between policy parameters. A small change in Euclidean distance between policy parameters might not be reflective of the change in the distributions. We can replace the Euclidean distance with a KL divergence between the policies, $\text{KL}(\pi_\pparams || \pi_{\pparams_t}) \doteq \int_\mathcal{A} \pi_\pparams(a) \ln \tfrac{\pi_\pparams(a)}{\pi_{\pparams_t}(a)} \mathrm{d}a$, giving instead\footnote{This exact formula is only true for $J(\pparams) \doteq \mathbb{E}_{\pi_\pparams}[\trueq(A)]$; it is slightly different for the mirror descent variants of other objectives $J$ for actor-critic methods. This is thoroughly described in a recent work \citep{neumann2025investigating}. We leverage the derivations from that work, but keep the explanation here simpler.}
	\begin{align*}
		\hat{J}_{\text{KL}}(\pparams) &\doteq J(\pparams_t) + (\pparams-\pparams_t)^\top \nabla J(\pparams_t) - \frac{1}{2 \eta}  \text{KL}(\pi_\pparams || \pi_{\pparams_t})\\
		\pparams_{t+1} &= \arg\max_{\pparams} \hat{J}_{\text{KL}}(\pparams) 
	\end{align*}
	This argmax no longer has a closed-form solution; instead, we use multiple steps of ascent on $\hat{J}_{\text{KL}}(\pparams)$.

	Many algorithms have been shown \citep{vieillard2020leverage} to use updates that are more similar to mirror ascent than to gradient ascent. These include TRPO \citep{trpo-schulman15}, MDPO \citep{tomar2022mirror}, MPO \citep{abdolmaleki2018maximum}, MD-MPI \citep{vieillard2020leverage}, PPO \citep{schulman2017proximal} and FMA-PG \citep{Vaswani22a-GeneralSurrogateRL} and variants \citep{neumann2025investigating}. 
	TRPO was originally motivated as a trust-region method, inspired by Conservative Policy Iteration (CPI) \citep{Kakade02}, which had guaranteed improvement if the policy did not change too much after each greedification step. 
	This update used a KL divergence to the previous policy and spurred a line of work further developing such algorithms, ultimately with unifications highlighting the connection to mirror ascent. For that reason, this update has taken a variety of forms. 
	We include the mirror-descent variant of SAC in Alg. \ref{alg:sac}, leveraging recent work that developed such a mirror-descent version \citep{neumann2025investigating}. 
	
	
	\subsection*{Actor Objectives using the KL}
	
	It is also possible to use different objectives for approximate greedification, namely the actor update. This change is, in fact, what primarily distinguishes SAC, MPO, and GreedyAC from AC(0). The key idea behind these methods is to choose a target policy $\pi_{\text{target}}$, and then the approximate greedification step involves reducing the KL divergence to that target policy. As was recently shown \citep{chan2021-greedification}, SAC and many other actor-critic methods can be seen as minimizing a mode-seeking KL divergence to the Boltzmann policy: 
	\begin{equation*}
		\text{KL}(\pi_\pparams || \pi_{\text{ent}}) \quad\text{  for } 
		\pi_{\text{ent}}(a) \propto \exp(q_w(a)/\beta)
	\end{equation*}
	for entropy parameter $\beta$. This Boltzmann policy is the solution to the entropy regularized greedification step: 
	\begin{equation*}
		\pi_{\text{ent}} \doteq \arg\max_{\pi} \mathbb{E}_\pi[q_w(A)] + \beta \mathcal{H}(\pi)
	\end{equation*} 
where  $\mathcal{H}(\pi) = \mathbb{E}_\pi[-\log\pi(A)]$ is the entropy of policy. As $\beta$ approaches zero, entropy regularization is removed, and we recover the greedy policy. Note that to optimize this objective, we only need to sample from $\pi_\pparams$, and never explicitly form the policy $\pi_{\text{ent}}$. 

This KL objective was introduced for the Soft Actor-Critic (SAC) \citep{haarnoja-SAC2018}, but many other algorithms adopt a similar loss in their approximate greedification step \citep[Table 2]{chan2021-greedification}. In fact, even the standard AC update can be seen as an instance of this update, because as $\beta$ approaches zero, the negative gradient of this KL objective approaches\footnote{Expanding the KL gives $-\nabla_\theta\text{KL}(\pi_\pparams || \pi_{\text{ent}}) = \tfrac{1}{\beta}\nabla_\pparams \mathbb{E}_{\pi_\pparams}[q_w(A)] - \nabla_\pparams \mathbb{E}_{\pi_\pparams}[\log{\pi_\pparams}(A)]$, and as $\beta \to 0$, the first term dominates and the update direction aligns with the standard policy gradient (AC) update $\mathbb{E}_{\pi_\pparams}[q_w(A)]$.} $\tfrac{1}{\beta}\nabla_\pparams \mathbb{E}_{\pi_\pparams}[q_w(A)]$. 
	
	Most methods use the mode-seeking KL written above, but there are a handful that use the mass-covering KL \citep[Appendix A3-A4]{chan2021-greedification}. This distinction comes from the fact that the KL is not symmetric: $D_{\text{KL}}(\pi || \pi_\theta) \neq D_{\text{KL}}(\pi_\theta || \pi)$, resulting in different preferences on the solution, as depicted in Figure \ref{fig:hero} (D). 
	The forward KL $D_{\text{KL}}(\pi || \pi_\theta)$ is mass-covering, whereas the reverse KL $D_{\text{KL}}(\pi_\theta || \pi)$ is mode-seeking.
	Consider that we want to approximate a multi-modal Boltzmann policy using a unimodal Gaussian. 
	One must weigh between covering the most significant mode of the Boltzmann or covering as much of the probability mass as possible.
	This is depicted in Figure \ref{fig:hero} (D). In practice, mass covering can cause the policy to more often select actions that do not correspond to any mode of a multimodal policy, resulting in poor performance \citep{Jiamin2026-revisitMixture}.

	It is also possible to pick other target policies and so get other updates. In MPO, the target policy is chosen to be 
	\begin{equation*}
		\pi_{\text{kl}} \doteq \arg\max_{\pi} \mathbb{E}_\pi[q_w(A)] - \tau \text{KL}(\pi || \pi_t)
	\end{equation*}
	for $\pi_t$ the previous policy, with a closed-form solution $\pi_{\text{kl}}(a) \propto \pi_t(a) \exp(q_w(a) / \tau)$. 
 		The loss for MPO uses a mass-covering KL loss to target policy $\pi_{\text{kl}}$. 
        GreedyAC uses a mass-covering KL to a target policy they call a percentile policy $\pi_\rho$.
        The algorithm maintains a slower-changing proposal policy $\tilde{\pi}$, samples several actions from $\tilde{\pi}$, and only keeps the top $\rho$ percentile. It increases the likelihood of only that top percentile of actions, for both $\pi_\pparams$ and $\tilde{\pi}$. They show that this approach is equivalent to reducing the KL to a percentile policy $\pi_\rho$ that shifts the action distribution in $\tilde{\pi}$ to put all weight on only the top percentile of actions. Because both MPO and GreedyAC use a mass-covering KL, the algorithms have a similar flavor: sample from a proposal policy (not from $\pi_\pparams$) and increase the likelihood of those actions.	
	
	\subsection*{Policy Parameterization}
	The policy $\pi_\pparams(a)$ is a distribution that can be parameterized in many different ways, which can significantly impact learning behavior. One of the most common choices is to use a Gaussian distribution, with some form of squashing or clipping to keep the actions within a bounded range. However, other densities have been considered, including beta \citep{chou2017improving} and Student's $t$ \citep{kobayashi2019student,Zhu2025-qExpPolicy}, as depicted in Figure \ref{fig:hero} (C). One could consider discretizing the continuous action space to use the softmax policy. 
	Recently, there has been more investigation into more expressive policies, including mixture models \citep{Jiamin2026-revisitMixture} and diffusion policies \citep{yang2023policy,psenka2023learning,ding2024diffusion}.
	
	The best-performing choice typically depends on the problem and other algorithmic components. Understanding the impact of the choice remains largely unresolved. There are several criteria to consider when making the choice. The actor slowly greedifies around high-valued actions according to the critic, but also has a role to provide sufficient exploration.\footnote{These two roles can be separated, but the most common choice is for the actor to be the same for the action selected in the bootstrap target and for behavior}
	All of these parameterizations introduce stochasticity to promote exploration, initialized with higher variance, with learning typically decreasing stochasticity while enforcing a lower bound on variance. Even deterministic policies, such as in DDPG \citep{Lillicrap2016-ddpg} and TD3 \citep{fujimoto2018addressing}, add stochastic (Gaussian) noise for exploration. In fact, this approach of adding Gaussian noise with variance $\sigma$ to a deterministic policy is equivalent to using a Gaussian parameterization in which the variance is fixed at $\sigma$ and not learned. It is not clear if more expressive policies, like diffusion policies, facilitate either of the two goals of sufficient exploration and 
	greedification. However, some evidence suggests they provide minimal to no advantages in the case of mixture policies \citep{Jiamin2026-revisitMixture} and unclear evidence in the case of diffusion policies~\citep{jain2025sampling}. In this work, we focus on comparing these simpler parameterized policies, including variants of the Gaussian, beta, and Student's $t$ distributions; see SI Appendix \ref{app:params} for detailed descriptions.
	
	\subsection*{Update-to-data Ratios}
	
	The update-to-date (UTD) ratio is the number of updates used per environment step.~\footnote{In online RL, the replay ratio, i.e., how often we sample a mini-batch per environment step, is used to discuss UTD. The concept is the same, but UTD is more general.} If the UTD for the actor is 10, the actor network is updated 10 times on every environment step, on average. A higher UTD can make the agent more sample-efficient, but care must be taken to avoid overfitting to the data seen so far. As the dataset grows, this issue becomes less of a concern during interaction, but in early learning, a high UTD could cause problems. This has been observed as a phenomenon called \textit{primacy bias} \citep{Nikishin2022-primacyBias}, and common wisdom is that it can be difficult to raise the UTD in reinforcement learning without careful adjustments to the architecture and hyperparameters~\citep{maheshwari2025addressing,d2022sample,voelcker2410mad}.  

	Furthermore, there is an interaction between the UTD for the actor and critic. If the UTD is higher for the critic than for the actor, we take more policy evaluation steps to obtain a more accurate critic estimate before greedifying. If the UTD is higher for the actor than the critic, then the actor is likely to be more greedy, and the critic evaluates a greedier policy. The seminal actor-critic theory assumes the first scenario: that the critic updates more quickly than the actor \citep{borkar1997stochastic,konda1999actor,Bhatnagar2009-NatureActorCritic}.

	\subsection*{Gradient Estimators}
	A key distinction in actor updates is how the policy gradient $\nabla_\theta J(\theta)$  is estimated.
	The likelihood-ratio (LR or score function estimator) and pathwise (PW or reparameterized estimator) are two common unbiased Monte Carlo estimators of the gradient  \citep{glynn1990likelihood,kingma2013auto}.
	The LR gradient estimator computes the gradient of 
	$q_w(A)$
	using the log-derivative trick:
	\begin{align*}
		\nabla_\theta J(\theta) &= \nabla_\theta \mathbb{E}_{A \sim \pi_\pparams}[q_w(A) ]  
		= \mathbb{E}_{A \sim \pi_\pparams}[q_w(A)  \nabla_\pparams \ln \pi_\pparams(A)]. 
	\end{align*}
	Adding in the baseline, $q_w(A) - v$, yields the familiar AC update.
	This estimator can have a high variance, but it can be applied to any differentiable policy distribution, including discrete ones.

    The PW estimator can be used when the policy allows a differentiable reparameterization.
	This is achieved by decoupling the stochasticity in action selection from the policy. For example, if the policy is a univariate Gaussian, 
    we can sample $\epsilon\sim\mathcal{N}(0,1)$ to produce the action: $a_t = g_\theta(\epsilon) = \mu_\theta + \sigma_\theta\epsilon$. In this setting, we can rewrite the gradient applying the chain rule directly through the critic and into the actor:
	\begin{align*}
		\nabla_\theta J(\theta) &= 
		\mathbb{E}_{\epsilon \sim p}[\nabla_\theta q_w(g_\theta(\epsilon))]\\
		&= \mathbb{E}_{\epsilon \sim p}[\nabla_A q_w(A) \nabla_\theta g_\theta(\epsilon)] \quad A = g_\theta(\epsilon),
	\end{align*}
	where $\epsilon \sim p$ is the noise sampled from a distribution independent of $\pparams$.
	This estimator enables end-to-end backpropagation through the sampled actions and often yields a lower-variance estimate. 
	However, this requires that $q_w(A)$ be differentiable with respect to $A$.
	Additionally, PW is not directly applicable to discrete distributions unless we use a continuous relaxation \citep{jang2016categorical}.
    
	Neither gradient estimator is strictly better than the other; each has its own advantages.
	Although PW is often shown to have lower variance than LR \citep{xu2019variance}, this is not guaranteed. 
	For example, it has been shown that a long chain of non-linear computations in model-based RL can cause the PW gradients to explode, whereas LR gradients remain robust \citep{parmas2018pipps}.
	
		\begin{figure}[t!]
		\centering
		\includegraphics[width=0.96\linewidth]{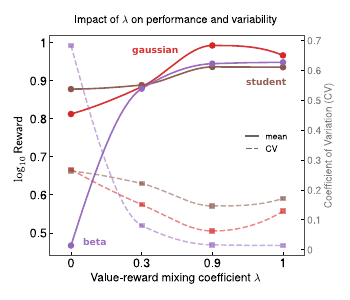}
		\caption{
			\textbf{Mixing immediate reward with value estimates improves stability and performance}
			For each policy parameterization, we fix all the hyperparameters to their default values (Table \ref{tab:common-hparams}) and vary the value reward mixing coefficient $\lambda \in \{0, 0.3, 0.9, 1\}$ in AC($\lambda$). 
			The actor update weights the gradient $\nabla \ln \pi_{\pparams}(A_t)$ by $\lambda R_t + (1-\lambda) q_w(A_t) - v$, interpolating between the critic's action-value estimate ($\lambda=0$) and the observed reward ($\lambda=1$). 
			Solid lines show mean reward (higher is better, left axis) averaged over 10 seeds while the dashed lines show coefficient of variation ($\mathrm{CV} = \sigma / | \mu |$, lower is better, right axis). 
			The curves are drawn using a shape-preserving cubic interpolation without altering pointwise values.
			Intermediate values of $\lambda$ achieve the best combined performance and variability across all policy parameterizations.
		}
        \vspace{0.5em}
		\label{fig:lambda}
	\end{figure}
    
	The LR estimator can always be used, whereas the PW estimator is restricted to algorithms that learn $q_w$ with a mode-seeking KL. 
	REINFORCE and PPO use the LR gradient typically with an advantage estimator, importance weighting and clipping.
	DDPG and TD3 can be viewed as deterministic special cases of the PW gradient.
	SAC typically uses PW in continuous action spaces and LR in discrete ones. 
	MPO and GreedyAC use the mass-covering KL, where the expectation is wrt to a proposal policy rather than $\pi_\theta$, side-stepping this issue. However, the updates for both of these algorithms can be best interpreted as an LR update as they apply a log-likelihood update on the sampled actions. 

	\section*{Results}
	In this section, we evaluate both the performance and stability of various component choices.
	The performance is defined as the mean reward across seeds (higher is better), whereas the stability refers to the run-to-run variability, measured by the coefficient of variation across seeds (lower is better).
	A reliable method should do well on both metrics. All the algorithms are tested on the Backwashing PID environment, constructed from data from a real water treatment system, depicted in Fig. \ref{fig:hero} (A), with a detailed description of the construction in the Materials and Methods section. Unless otherwise stated, all experiments use the adaptive critic UTD ratios as described in Alg. \ref{alg:critic}. We present summarized conclusions in the main text, and additionally provide detailed run-to-run and hyperparameter variability analysis in the Supplement \ref{app:extended_results}.
    
	\begin{figure}[t!]
		\centering
		\includegraphics[width=0.96\linewidth]{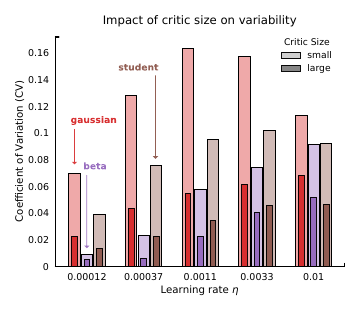}
		\caption{
			\textbf{Increasing the critic size reduces run-to-run variability	}.
			For each policy parameterization and learning rate, we compare AC($\lambda$) with a small critic (2 hidden layers of 64 units) and a large critic (2 hidden layers of 256 units), with all the hyperparameters fixed at their default values (Table \ref{tab:common-hparams}) and $\lambda=0$.
			The bar height represents the coefficient of variation (CV) across $N=10$ seeds, where lower values indicate more stable runs.
			Larger critics consistently reduce CV across the entire learning-rate range for all policy parameterizations.
		}
		\label{fig:critic_size}
	\end{figure}	
    	\begin{figure}[t!]
		\centering
		\includegraphics[width=0.96\linewidth]{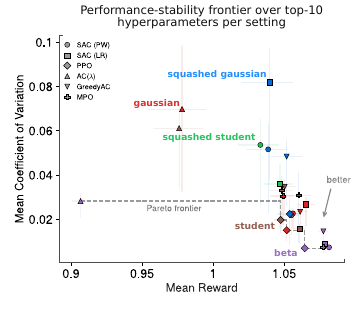}
		\caption{
			\textbf{Performance-stability frontier across algorithms and policy parameterizations}.
			For each \{algorithm, policy\} combination, we select the top-10 hyperparameter configurations by mean reward and aggregate over their 100 runs (10 configurations x 10 seeds). 
			The x-axis shows the mean reward (higher is better) averaged across these runs.
			The y-axis shows the mean coefficient of variation (lower is better), computed by taking the CV across seeds within each configuration and then averaging across 10 configurations: ($\mathrm{CV}_\text{mean} = \frac{1}{N}\sum_i \mathrm{CV}_i$ with $N=10$).
			This separates within-configuration variability (run-to-run noise at fixed hyperparameters) from cross-configuration variability (sensitivity to hyperparameter choice). 
			Error bars are 95\% bootstrap confidence intervals (n=2000 resamples).
			Marker shape indicates the algorithm, and color indicates the policy parameterization.
			The dashed Pareto frontier connects configurations for which no other plotted point achieves both higher mean reward and lower mean CV, and is computed directly from the plotted aggregate points.
			The best region of the frontier, with high mean reward and low variability, consists entirely of beta configurations, with five of the six algorithms clustering together.
			Student's $t$ and Gaussian appear on the frontier only at intermediate performance levels, while squashed policies are strictly dominated by multiple other configurations.
			Additionally, unlike other algorithms, which are spread apart, PPO configurations are relatively similar in performance across multiple policies.
		}
		\label{fig:pareto}
	\end{figure}
    \vspace{0.5em}
	\begin{ourconclusion}
		Mixing immediate reward with value estimates ($\lambda > 0$) improves stability and performance.
	\end{ourconclusion}
	\noindent AC($\lambda$) interpolates between the critic's action-value estimate ($\lambda=0$) and the observed reward ($\lambda=1$) in the policy gradient.
	In our experiments, relying entirely on the critic ($\lambda=0$) led to high run-to-run variability across all policy parameterizations, with beta policies being particularly affected. 
	This is likely due to value misspecification, introducing bias into the policy update.
	Introducing even a moderate reward mixing ($\lambda=0.3$) recovers most of the performance gap while significantly reducing variability, with the best effect consistently observed at $\lambda=0.9$.
	Completely removing the critic contribution ($\lambda=1$) tends to either reduce performance or increase variability for Gaussian and Student's $t$ policies.
	See Fig. \ref{fig:lambda} for the effect of $\lambda$ on performance and variability and Fig. \ref{fig:acl_all} for a more detailed breakdown across different hyperparameter and environmental noise configurations.

	\begin{figure*}[t!]
		\centering
		\includegraphics[width=0.99\linewidth]{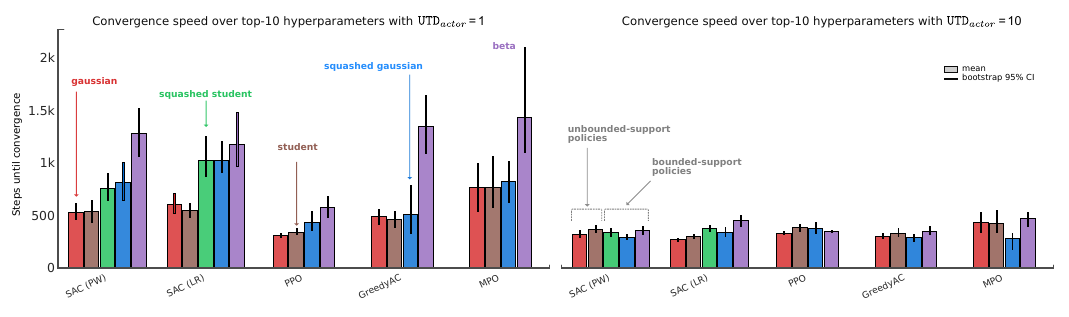}
		\caption{
			\textbf{Bounded-support policies converge slower than unbounded-support across algorithms}.
			For each \{algorithm, policy\} combination, we report the number of steps until convergence, defined as the first step at which the return reaches within 90\% of the policy's own final 5\% average performance.
			Note that this is a within-policy convergence metric, as the convergence does not guarantee reaching some predefined global threshold.
			Results are aggregated by pooling all runs from the top-10 hyperparameter configurations for each combination, then computing the mean (height of the lighter bar) and 95\% bootstrap confidence intervals (dark bars, n=2000 resamples).
			\textbf{Left:} at default actor UTD ($\texttt{UTD}_{actor} = 1$), the policies with bounded support converge slower than unbounded support in most cases, with the beta policy often requiring 2-3x more steps than Gaussian.
			\textbf{Right:} at $\texttt{UTD}_{actor} = 10$, this ordering largely collapses and all policies converge at comparable rates.
		}
                \vspace{0.5em}
		\label{fig:pareto_cv_r}
	\end{figure*}
		
	\begin{figure*}[t!]
		\centering
		\includegraphics[width=0.99\linewidth]{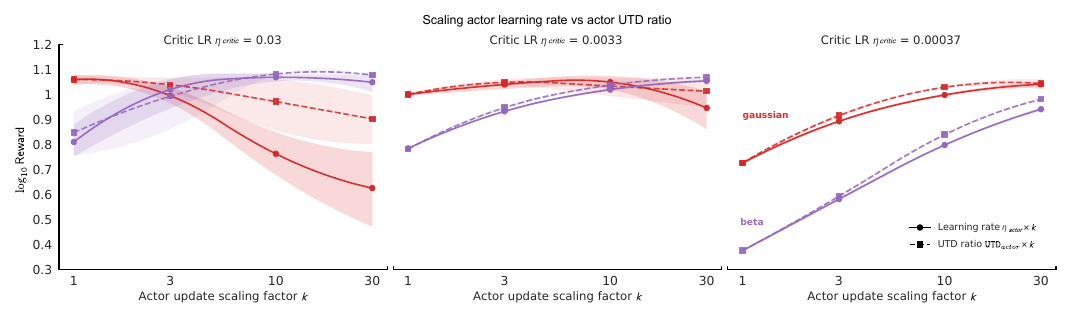}
		\caption{
			\textbf{Actor update-to-data ratio is a safer tuning knob compared to the actor learning rate.}
			For each policy (Gaussian, beta), we compare two strategies for scaling the actor update magnitude by a factor $k \in \{1,3,10,30\}$ i.e. scaling the actor learning rate $\eta_{actor} \times k$ (solid lines) or scaling the actor UTD ratio $\texttt{UTD}_{actor} \times k$ (dashed lines), across three representative critic learning rates $\eta_{critic}$ (different panels) and default SAC hyperparameters (Tables \ref{tab:common-hparams}, \ref{tab:algo-specific-hparams}).
			Before scaling is applied, the actor and critic learning rates are equivalent, i.e., $\eta_{actor} = \eta_{critic}$.
			Results are shown as mean reward across 10 seeds, and the shaded bands represent 95\% bootstrap confidence intervals with n=1000 resamples.
			For beta policies, the dashed and solid lines overlap across all learning rates, indicating that the two scaling strategies are almost identical.
			For Gaussian policies, they are equivalent to low critic learning rates but start diverging at high $\eta_{critic}=0.03$ (left), where scaling actor LR causes collapse at $k \geq 10$ while scaling actor UTD remains relatively stable.
			This suggests that the Gaussian policy's instability at high actor LR arises due to big updates to both the actor and critic networks at the same time, as opposed to large actor parameter movements alone.
		}

		\label{fig:utd_lr}
	\end{figure*}

\begin{ourconclusion}
	Critic inaccuracy is a major contributor to instability.
\end{ourconclusion}
Inaccurate critics increase run-to-run variability: some runs achieve good performance, whereas others fail, thereby reducing stability.  
There are several lines of evidence supporting this conclusion. Firstly, increasing $\lambda$ in AC($\lambda$) reduces the critic's contribution to the policy gradient.
Fig. \ref{fig:lambda} shows that higher $\lambda$ values decrease variability across all policy parameterizations. 
Secondly, increasing the critic size increases the critic's representational capacity, 
reducing critic error as long as the critic is trained sufficiently to exploit this added capacity.
Because our adaptive critic update (Alg.~\ref{alg:critic}) trains the critic until its batch error falls below a fixed threshold, the larger network actually realizes this lower error.
Consistent with this, Fig.~\ref{fig:critic_size} shows that larger critic networks reduce variability across all policy parameterizations in AC($\lambda$).
Finally, note that PPO in our non-contextual bandit setting does not require learning an action-value critic and, correspondingly, its configurations are among the most favorable in the performance-stability trade-off. Three of the four PPO configurations are Pareto efficient in Fig. \ref{fig:pareto} while the remaining one lies close to the Pareto frontier.

 \begin{ourconclusion}
		Beta policies achieve the best performance-stability tradeoff.
	\end{ourconclusion}
	\noindent 
    We aggregate results over the top-10 hyperparameters for each algorithm, including allowing different UTD per algorithm. In Fig. \ref{fig:pareto} we observe two patterns.
	Firstly, the best region of the performance-stability frontier consists entirely of beta configurations, with five different algorithms clustering tightly around near-optimal performance.
	This region denotes configurations that have high mean reward and low variability across runs and hyperparameters.
	Secondly, except for AC($\lambda$), no non-beta policy parameterization dominates any beta configuration. 
	The variants of AC($\lambda$) that perform worse are likely due to not sweeping over the actor UTD ratio in that setting.
	Some Student's $t$ and Gaussian configurations also lie on the frontier, but only in intermediate regions where they achieve lower mean reward but competitive variability. 
	In contrast, squashed Gaussian, which is the widely-used default in SAC implementations, performs poorly. 
	For a more detailed breakdown, see SAC with PW estimator, SAC with LR estimator, PPO, AC($\lambda$), GreedyAC in Figs. \ref{fig:sac_reparam}, \ref{fig:sac_noreparam}, \ref{fig:ppo_utd}, \ref{fig:acl_all} and \ref{fig:greedyac} respectively.

	\begin{figure*}[t!]
		\centering
		\begin{minipage}{0.99\textwidth}
			\includegraphics[width=0.99\linewidth]{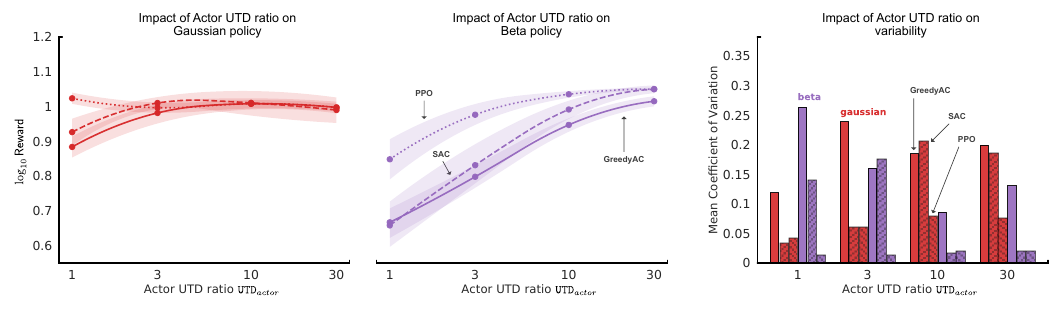}
		\end{minipage}
		\caption{ 
			\textbf{Higher actor update-to-data ratio improves beta policies but destabilizes Gaussian policies.}
			We computed results using top-5 learning rates for every $\{\text{policy}-\texttt{UTD}_{actor}\}$ combination while fixing the other hyperparameters to their default values from Table \ref{tab:algo-specific-hparams}.
			\textbf{Left and middle:} Mean rewards obtained by GreedyAC, SAC and PPO when changing the actor UTD, where higher is better. Each dot is the average over all selected runs. 
			The shaded area represents 95\% bootstrap confidence interval over 50 runs with n=2000 resamples.
			The curves are drawn using a quadratic B-spline interpolation for visual clarity without altering pointwise values.
			With a Gaussian policy, the impact of changing $\texttt{UTD}_{actor}$ on the average performance is minimal: PPO and GreedyAC decrease, whereas SAC increases.
			On the other hand, the average performance of the beta policy improves very significantly as  $\texttt{UTD}_{actor}$ is increased. 
			\textbf{Right:} Mean variability for different Actor UTD ratios, where lower is better. 
			Since we are aggregating results over multiple hyperparameter configurations, we compute the variability ($\mathrm{CV}_{i} = \sigma_i / | \mu_i |$) separately for each (top-5) hyperparameter configuration and average over these variabilities to give us a single number  ($\mathrm{CV}_\text{mean} = \frac{1}{N}\sum_i \mathrm{CV}_i$ where $N=10$ configurations).
			As we increase $\texttt{UTD}_{actor}$, the variability increases for Gaussian and decreases for beta policy.
			With the best hyperparameter configurations, the beta policy achieves higher mean reward and lower mean variability than the Gaussian policy across all three learning algorithms.
		}
                \vspace{0.5em}
		\label{fig:actor_utd}
	\end{figure*}
\begin{figure*}[t]
		\centering
		\begin{minipage}{0.99\textwidth}
			\includegraphics[width=0.99\linewidth]{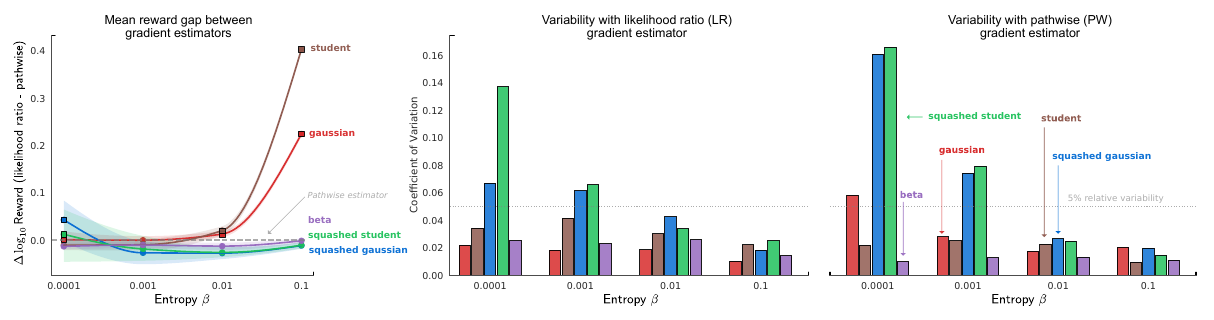}
		\end{minipage}
		\caption{
			\textbf{Pathwise gradient estimators are often unreliable with clipped and squashed policy parameterizations.}
			We compare pathwise (PW / reparameterized) and likelihood ratio (LR / non-reparameterized) gradient estimators in Soft Actor-Critic by picking the same hyperparameters for each policy parameterization, and display their performance across different entropy values $\beta$. \textbf{Left:} absolute difference in mean reward obtained between the two gradient estimators, computed as $\Delta \log_{10}(\text{Reward}) = \log_{10}(\text{Reward}_{\text{LR}}) - \log_{10}(\text{Reward}_{\text{PW}})$.
			The curves are drawn using a shape-preserving cubic interpolation to improve visual clarity without altering the pointwise values.
			The dashed \textit{baseline} line represents the mean performance of the reparameterization trick variant.
			The shaded region represents the bootstrap confidence intervals based on n=2000 resamples.
			A point above the dashed line implies that the corresponding setting has a higher mean performance when used with an LR estimator.
			We observe that the policies with clipped sampling (Student's $t$ and Gaussian) obtain a higher mean reward with an LR estimator, whereas the policies with natural boundedness (beta) and squashed sampling (squashed Gaussian) behave similarly under both estimators. 
			\textbf{Middle and right:} coefficient of variation ($\mathrm{CV} = \sigma / | \mu |$) of the total reward across training runs for the LR and PW versions, respectively. Bars show the relative variability across all policies at different entropy values, with smaller values indicating a more stable setting. The squashed Gaussian policies are the least stable under both estimators, with a very high variability when used with the PW estimator.}
		\label{fig:reparam_vs_log}
	\end{figure*}

	\begin{ourconclusion}
		Bounded-support policy parameterizations are slower to converge compared to unbounded-support ones.
	\end{ourconclusion}
	\noindent 
	We group the squashed and beta policy parameterizations as bounded-support, and the clipped Gaussian and Student's $t$ as unbounded-support.
	From Fig. \ref{fig:pareto_cv_r} (middle), we observe that the effect depends on how strongly the action-space constraints are enforced by the policy parameterization.
	With the default actor UTD ratio of 1, the beta policy consistently learns the slowest among all algorithms, followed by squashed policies and then clipped policies. 
	The clipped policies are based on unbounded densities and therefore can be optimized in a relatively unconstrained parameter space.
	The squashed policies enforce bounds through a smooth non-linear transformation, which introduces additional curvature and sensitivity near the boundaries.
	In contrast, the beta policy enforces bounds through its density, and its parameters couple together to control the location, shape, and skewness of the distribution simultaneously, making optimization more difficult.
	Once we increase the actor UTD ratio (Fig. \ref{fig:pareto_cv_r} (right)), the differences in convergence speed among the various policy parameterizations largely disappear.
	However, if the critic is not sufficiently accurate, a high actor UTD could amplify the critic's errors and steer the policy towards poor solutions.

	\begin{figure*}[t!]
		\centering
		\begin{minipage}{0.99\textwidth}
			\includegraphics[width=0.99\linewidth]{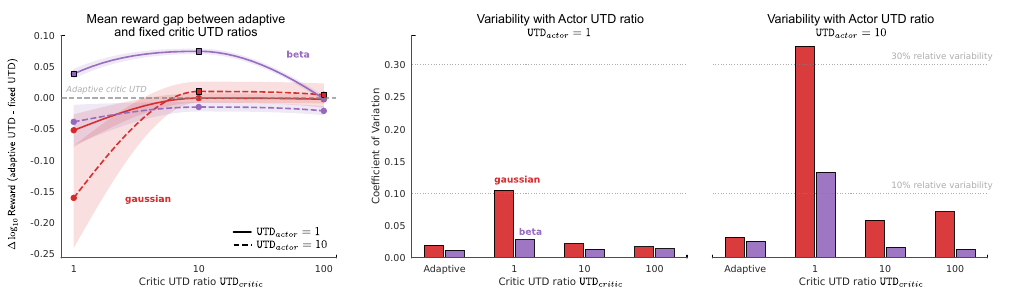}
		\end{minipage}
		\caption{\textbf{Adaptive critic updating is more reliable than a fixed critic update-to-data ratio.}
			For each combination of policy, actor UTD ratio $\texttt{UTD}_{actor}$ and critic UTD ratio $\texttt{UTD}_{critic}$, we select the best-performing learning rate and aggregate over its 10 seeds.
			\textbf{Left:} Reward gap between the fixed critic UTD ratio $\texttt{UTD}_{critic}$ and adaptive critic UTD ratio, computed as $\Delta \log_{10}(\text{Reward}) = \log_{10}(\text{Reward}_{\mathrm{fixed}}) - \log_{10}(\text{Reward}_{\mathrm{adaptive}})$, so values above the dashed reference line at zero (adaptive UTD performance) indicate that the fixed ratio outperforms adaptive, and values below indicate the opposite.
			Solid and dashed curves correspond to $\texttt{UTD}_{actor}=1$ and $\texttt{UTD}_{actor}=10$, respectively.
			Shaded bands represent 95\% bootstrap confidence intervals over 50 runs with n=2000 resamples, and curves use shape-preserving cubic interpolation between measured points for visual clarity.
			\textbf{Middle and right:} coefficient of variation for $\texttt{UTD}_{actor}=1$ and $\texttt{UTD}_{actor}=10$, respectively, with adaptive UTD shown as the leftmost pair of bars in each panel for direct comparison. Horizontal dotted lines mark 10\% and 30\% relative variability as fixed reference points to ease comparison across panels and figures.
			Fixed critic UTD ratios are competitive with adaptive only for the beta policy at low actor UTD $\texttt{UTD}_{actor}=1$. In every other configuration, fixed ratios either yield lower mean reward (left panel, points below baseline) or substantially higher variability (middle and right panels). The Gaussian policy at $\texttt{UTD}_{critic}=1$ is the most extreme failure mode, with variability rising to roughly 11× the adaptive value when $\texttt{UTD}_{actor}=10$.}
		\label{fig:critic_utd}
	\end{figure*}
    
	\begin{ourconclusion}
		Actor update-to-date ratio (UTD) is a safer tuning knob compared to the actor learning rate.
	\end{ourconclusion}
	\noindent 
	Scaling the actor learning rate $\eta_{actor}$ by $k$ and the actor $\texttt{UTD}_{actor}$ by $k$ are both ways to control how much the parameters are updated. 
    We find scaling either has a very similar effect for the beta policy across the three critic learning rates $\eta_{critic}$ depicted in Figure \ref{fig:utd_lr}.
	For a Gaussian policy, however, the behavior is different at high critic learning rates: increasing the actor learning rate produces a significantly stronger performance collapse compared to increasing the actor UTD ratio.
	This suggests that the Gaussian policy's instability stems from simultaneous fast updates to both the critic and actor networks, rather than from actor parameter movements alone.
	This indicates that the actor UTD ratio is a safer tuning knob.

    \begin{ourconclusion}
		Higher actor update-to-data ratio improves beta policies but destabilizes Gaussian policies.
	\end{ourconclusion}
	\noindent 
    $\texttt{UTD}_{actor}$ has different effects depending on the policy parameterization (Fig. \ref{fig:actor_utd}).
	For beta policies, increasing $\texttt{UTD}_{actor}$ from 1 to 30 significantly improves the mean reward and lowers run-to-run variability consistently across PPO, SAC and GreedyAC.
	In contrast, for Gaussian policies, the mean reward changes minimally while the variability increases significantly, rising by a factor of four for SAC and roughly doubling for GreedyAC and PPO.
    
	\begin{ourconclusion}
		Pathwise gradient estimators are often unreliable with clipped or squashed policies.
	\end{ourconclusion}
	\noindent 
	Most continuous-control codebases use the pathwise (PW) gradient estimator by default, often without exposing it as a tunable choice. 
	Whether this is a good default, however, depends on the policy parameterization.
	PW gradients flow through the sampled action, so any transformation that distorts this path such as clipping or squashing can negatively affect the gradient signal.
	In contrast, the likelihood ratio (LR) gradients are unaffected by this issue, since they do not differentiate through the sampled action.
    
	Fig. \ref{fig:reparam_vs_log} shows three regimes. 
	Firstly, clipped policies (Gaussian and Student's $t$) suffer a significant drop in performance at higher entropies under PW, but remain robust under LR. 
	This is likely because the high entropy pushes more sampled actions to the clipping boundary, where they contribute zero PW gradient.
	Secondly, squashed policies achieve similar mean rewards under both estimators, but a lower run-to-run variability under the LR estimator.
	Finally, the beta policy is essentially insensitive to the choice of estimator in both mean performance and variability, likely because its bounds are encoded in the density rather than controlled by a transformation, so there is no clipped or squashed path for the gradient to flow through.
	The practical implication is to treat PW as a choice rather than a default whenever the policy involves clipped or squashing, and to consider LR in those settings.

	\begin{ourconclusion}
		Adaptive critic updating is more reliable than a fixed critic update-to-data ratio.
	\end{ourconclusion}
	\noindent 
	Comparing the adaptive critic UTD ratio (Alg. \ref{alg:critic}) with the commonly used fixed critic UTD strategy (Alg. \ref{alg:critic_fixed_utd}), we observe that adaptive updating largely avoids the failure modes of fixed critic UTD ratios (Fig. \ref{fig:critic_utd}).
	The worst case is $\texttt{UTD}_{critic}=1$, which produces 11x higher run-to-run variability for Gaussian policies and 5x for beta under $\texttt{UTD}_{actor}=10$, as well as lower mean reward across configurations.
	Higher fixed ratios of $\texttt{UTD}_{critic}=\{10,100\}$ perform comparably; however, they require the practitioner to either know or tune the appropriate value in advance.
	The one case where the fixed ratio can outperform adaptive is beta with $\texttt{UTD}_{actor}=1$; however, both perform poorly compared to higher $\texttt{UTD}_{actor}=10$. 
	
    Overall, we find adaptive updating 
    replaces the more sensitive $\texttt{UTD}_{critic}$ hyperparameter with an easier-to-interpret error threshold $\omega$, which can be set based on the magnitude of the critic loss.
	In principle, a threshold $\omega$ defined relative to a normalized reward scale should transfer across environments more naturally than a raw update count, whose appropriate value depends on batch size, critic's representational capacity, and the complexity of the environment.
	
	\section*{Discussion}
	We organize the discussion around common themes that emerge from these results and their implications for the full sequential reinforcement learning setting.
		
	\subsection*{Common empirical themes}
	The first theme is that critic accuracy controls actor-critic stability, and many of the differences in run-to-run variability we observed are connected to it.
	Mixing immediate reward into the policy gradient (AC($\lambda$), with $\lambda>0$) reduced the run-to-run variability across every policy parameterization we tested.
	A larger critic, with greater representational capacity to learn the value surface, reduced run-to-run variability across the entire learning-rate range.
	Adaptive critic updating, which keeps the critic error below a threshold before each actor update, outperformed every fixed update-to-data ratio we tried, except in one configuration that corresponded to a suboptimal agent.
	Most importantly, PPO, which in a bandit setting requires no learned critic, produced some of the most favorable points on the performance-stability frontier.
    
	The common reason is that when the critic is wrong, the actor inherits that error at every gradient step, and the resulting instability manifests more prominently as high run-to-run variability rather than low average reward.
	For deployment, this means that instead of only considering whether to use a specific algorithm as a whole, we should consider whether the chosen implementation has sufficient critic capacity, sufficient evaluation steps before each greedification, and a convenient way to trade off bootstrapped estimates against the observed signal.
	Adaptive critic updating, in particular, is a general method that can be added to most actor-critic implementations.
	
	The second theme is that policy parameterization is a dominant factor driving differences between agents. The same algorithm with two different policy parameterizations behaved more differently than two algorithms with the same parameterization.
	Gaussian and beta policies react oppositely in response to the actor UTD: increasing the actor UTD improved beta's mean reward and reduced its run-to-run variability, whereas the same change destabilized Gaussian policies. 
	Gradient estimator choice interacts similarly with parameterization: the clipped Gaussian and Student's $t$ policies suffered significant performance degradation at higher entropies under pathwise gradients but remained robust under likelihood-ratio gradients, whereas beta was insensitive to the choice of gradient estimator.
	The squashed Gaussian policy, as the standard default in continuous-control SAC setups, sat in a strictly dominated region of the performance-stability frontier and exhibited high run-to-run variability under pathwise gradients.
	The takeaway is that switching the policy parameterization within an algorithm can change the outcome more than switching the algorithm itself, a fact often overlooked in the standard practice of selecting an algorithm and adopting its default parameterization.
		
	Finally, many current default choices exhibit poor run-to-run variability, which is not easy to detect when only looking at average performance across runs. 
	The pathwise gradient estimator is the default in most continuous-control codebases, but can have high run-to-run variability under the standard choices of clipped and squashed policies. 
	Fixed critic UTD ratios are usually set to the values that worked for the algorithm's original set of benchmark environments, but this choice seems sensitive.
	The squashed Gaussian policy is a default for SAC, but it performed much worse than a beta policy.
	Each of these component-level decisions contributes significantly to the deployment reliability.
	
\subsection*{Implications beyond the bandit setting}
We can consider outcomes in the more general Markov decision process (MDPs) setting considered in reinforcement learning. In an MDP, the agent's actions affect a state that evolves over time, and the consequences of an action are distributed over a long horizon of future rewards rather than realized immediately. Moving to the general MDP setting introduces several new questions to investigate, such as the effect of bootstrapped targets in the critic, exploration operating over the state space rather than the action space alone, more significant distribution shifts in the replay as the policy improves, and the tighter coupling between actor and critic as now the actor impacts the critic's bootstrapped target.


Several conclusions in this paper are based on per-step mechanisms that should transfer to the MDP setting.
Clipped and squashed policy parameterizations having distorted gradients under pathwise estimators is a property of how a single sampled action contributes to a gradient step; states and trajectories do not impact this.
The same reasoning applies to the beta policy's robustness to estimator choice, which stems from its bounds being imposed by the density itself rather than by a transformation.
Adaptive critic updating can also be used in the MDP setting, and should have similar benefits. There could be new outcomes, however, because the measured error (the temporal difference error) is now a surrogate for the value error, unlike the bandit setting, where the squared reward error is a clear signal of critic accuracy. 

There are some findings that are likely to be more notably different in the MDP setting compared to the bandit setting. 
The optimal AC($\lambda$) mixing point in our experiments was around $\lambda=0.9$, which placed most of the weight on the observed reward rather than the critic. In MDPs, the alternative to a bootstrapped critic estimate is not a single-step reward but a multi-step or $\lambda$-return---equivalently a generalized advantage estimator in PPO---whose variance grows with the horizon and the discount factor. The mixing parameter trades off the bias of an inaccurate critic and the variance of an unbiased target. This trade-off is the same in MDPs, but the variance is much larger. Furthermore, if the data is off-policy, we need to incorporate importance-sampling ratios into the $\lambda$-return, which further increases variance. The use of $\lambda = 0$, as in SAC, MPO, and GreedyAC, is more typical when using buffers in the MDP setting, as this avoids these issues with off-policy $\lambda$-returns. In the MDP setting, on-policy algorithms like PPO do not leverage large replay buffers and are often less sample efficient than algorithms like SAC; we expect this sample efficiency difference to be more significant in the MDP setting than it was in our bandit setting. Further, in the MDP setting, changes to the actor impact the critic update by changing the bootstrap target, whereas in the bandit setting it only impacts the data that is gathered. With a higher actor UTD in MDPs, the critic now has to track a faster changing actor, and so we might find a more complex relationship between actor UTD and critic UTD. 

\subsection*{Future directions}
The most direct next step is to test which of our conclusions transfer to the standard MDP setting.
Each conclusion is falsifiable through a focused empirical study.
Another direction is to extend the analysis to non-stationary settings, which are common in many real-world control tasks. 
Industrial controllers must adapt as the equipment degrades, as the operating conditions change, and as the system's operating points change.
The component choices that dominate reliability in our stationary setting may not necessarily be the ones that dominate reliability under non-stationarity, and identifying which ones matter would be valuable in practice.

\definecolor{actionscol}{RGB}{0, 100, 255} 
\definecolor{internalcol}{RGB}{0, 255, 100} 
\definecolor{envcol}{RGB}{255, 0, 100} 
\newcommand{\cvar}[1]{\textcolor{actionscol}{#1}}
\newcommand{\cvarint}[1]{\textcolor{internalcol}{#1}}
\newcommand{\cvarenv}[1]{\textcolor{envcol}{#1}}
\matmethods{
	\subsection*{Evaluation Criteria}
What makes a good learning algorithm? We want an algorithm that achieves 1) High overall mean reward during deployment, 2) Low run-to-run variability, and 3) Low sensitivity\footnote{In RL, hyperparameter sensitivity typically is used to mean one of two things. An algorithm's hyperparameters are sensitive if: 1) good performance in two different environments requires tuning the hyperparameters for each, 2) Small changes to the hyperparameters cause significant changes to performance in one environment. In this work, we are concerned with \#2.} to different hyperparameters. Accordingly, we analyze performance distributions using the coefficient of variation ($\mathrm{CV} = \sigma / | \mu |$ where $\sigma$ is the standard deviation and $\mu$ is the mean of reward $r$) as a scale-invariant measure of run-to-run variability. 
A CV of 0.1 represents 10\% relative variability regardless of the reward scale, allowing direct comparisons across algorithms, hyperparameter configurations, and, in principle, across environments.
Additionally, we report bootstrap confidence intervals on aggregate statistics to quantify uncertainty.
The figures in the main text are presented as evidence for the conclusions and are summarized from the extended results in Appendix \ref{app:extended_results}.

\subsection*{Experimental Setup}
We conducted all experiments in our Backwashing-PID Environment.
Each hyperparameter configuration was trained using 10 random seeds for $T=5000$ environment interaction steps. At every step, we sampled a triplet of PID gains: $\{p, i, d\} \in [0, 20]$ from the policy, which were scaled appropriately depending on policy parameterization.  
Each experiment is swept over the entire range of learning rates for the actor and critic $\{0.00012, 0.00037, 0.0011, 0.0033, 0.01, 0.03, 0.09\}$, and both are optimized with Adam $(\beta_1, \beta_2) = (0.9, 0.999)$. 
For the main sweeps, we fix actor and critic learning rates to be equal to reduce the hyperparameter search space, and vary them independently in specific experiments (Fig \ref{fig:utd_lr}).
The critic is a 2-hidden-layer MLP with 64 units per layer.
Adaptive critic UTD follows Algorithm~\ref{alg:critic} with tolerance $\omega = 10^{-3}$ and a maximum of 100 updates per interaction step.
Replay buffer sampling begins after the first step; the mini-batch size is $\min(512,\, |\mathcal{B}|)$ where $|\mathcal{B}|$ is the current buffer size.
We rank the hyperparameters by their mean reward across the entire training run (area under the curve).
Since the rewards are normalized to $[-1,0]$, we report the performance as $-\log_{10}(|\text{Reward}|)$ so that higher values indicate better performance.
For brevity, figure axes display this as $\log_{10}(\text{Reward})$ without negation and absolute value.
See Table ~\ref{tab:exp-overview} for an overview of all the algorithm-specific hyperparameters considered under this study, Table~\ref{tab:env-hypers} for environment hyperparameters, and Tables~\ref{tab:common-hparams}, \ref{tab:algo-specific-hparams} for the fixed hyperparameters.

	\subsection*{Backwashing-PID Environment}
	
	We evaluate actor--critic methods on a real-world control task: tuning a PID controller \citep{astrom1995pid} that regulates water flow for backwashing for drinking water treatment. Our motivation is to explore algorithmic behavior across a range of component choices and hyperparameters in a realistic setting, since tuning controllers on live systems is costly and often impractical. To this end, we built a simulator by fitting a second-order polynomial to empirical data collected from the drinking water treatment plant in Drayton Valley, Alberta (pump speed $s$ and flow rate $f$). We will describe our simulator in this section. This is a continuing problem with no episodes or state resets.
	
	Our simulator is defined over discrete internal (hidden to the agent) timesteps $j \in \{1, \dots, \mathcal J\}$, corresponding to a single control cycle. Each of these internal timesteps represents 1 second of water flow through the pump. At each agent--environment interaction step $\cvarenv{t}$, the agent selects PID gains $\{\cvar{p}_{\cvarenv{t}}, \cvar{i}_{\cvarenv{t}}, \cvar{d}_{\cvarenv{t}}\}$, which are held constant for that control cycle. Then, the environment steps through the internal timesteps $j$ as follows:
	
	\begin{enumerate}
		\item \textsf{Flow generation:} Given current pump speed $s_j$ (with $s_0=20$), the simulator produces the noisy flow rate:
		\begin{align}
			\begin{split}
				f_j = 6.0553 \times 10^{-4} &\times s_j^2 - 2.530755 \times 10^{-2} \times s_j \\
				&+  0.26727398 + \varepsilon^{(f)}_j
			\end{split}
			\label{eq:flow_eq}
		\end{align} 
		
		where $\varepsilon^{(f)}_j \sim \mathcal{N}(0, \sigma_f^2)$ represents the measurement noise with the fixed standard deviation of $\sigma_f=0.003$ in all our experiments.
		
		\item \textsf{Error and PID control:} We compute the flow error $E_j = f_j - f_\text{ideal}$ and update the pump speed:
		\begin{align}
			s_{j+1} &= \cvar{p}_{\cvarenv{t}} E_j + \cvar{i}_{\cvarenv{t}} \sum_{k=1}^j E_k \Delta t + \cvar{d}_{\cvarenv{t}} \frac{E_j - E_{j-1}}{\Delta t} \\
			s_{j+1} &= \operatorname{clip}\left( s_j, s_{\min}, s_{\max} \right)
			\label{eq:pid_eq}
		\end{align}
		where $\Delta t = 1$ second is the time period between each internal timestep, $s_{\min} = 0$ is the minimum pump speed, $s_{\max} = 100$ the maximum pump speed, and $E_0=0$ is the initial error.

	\end{enumerate}

	After completing a control cycle, we can compute the reward $r_{\cvarenv {t}}$ based on the performance of the controller. The goal in this environment is to adjust the PID controller parameters such that the flow rate $f_j$ over the control cycle $1\dots \mathcal J$ matches the setpoint value $f_{ideal}$ as quickly and stably as possible. 
	To this end, we first define an asymmetric deviation term that penalizes overshooting more heavily than the undershooting:
	\begin{align}\label{eq:deviation}
		v_j = \begin{cases}
			5(f_j - f_{\mathrm{ideal}}), & \text{if } f_j > 1.05 f_{\mathrm{ideal}}, \\
			f_j - f_{\mathrm{ideal}}, & \text{otherwise}.
		\end{cases}
	\end{align}
	
	Using Eq. \ref{eq:deviation}, we compute the total deviation vector $\vec{v} = (v_1, \dots, v_{\mathcal J})$ and compute the reward for timestep $\cvarenv{t}$ as:
	\begin{align}
		\label{eq:reward_noise}
		r_{\cvarenv{t}} =  \min\!\left(0,\; -\frac{\|\vec{v}\|_2}{110} + \varepsilon^{(r)}_{\cvarenv{t}}\right), \quad \varepsilon^{(r)}_{\cvarenv{t}} \sim \mathcal{N}(0, \sigma_r^2)
	\end{align}
	
	where $\|\vec v\|_2$ denotes the Euclidean norm of the deviation vector, and the additive $ \varepsilon^{(r)}$ term models fixed environmental noise using $\sigma_r = 0.01$. The constant 110 and the clipping ensure that the rewards are roughly normalized to the $[-1,0]$ range. All key environment hyperparameters are summarized in Table~\ref{tab:env-hypers}. Fig. \ref{fig:pid_surfaces} illustrates the reward surfaces. 

\subsection*{Adaptive critic updates}
Unless otherwise stated, we use an adaptive critic UTD strategy (Alg. \ref{alg:critic}), 
rather than the commonly used fixed critic UTD strategy (Alg. \ref{alg:critic_fixed_utd}).
At each environment step, we perform critic updates until the batch error falls below the threshold $\omega$ or we reach the maximum number of updates $n_\text{steps}$.
This typically keeps the critic error below $\omega$, subject to the ceiling set by $n_\text{steps}$, before the critic is used to update the actor, thereby reducing one confounding source when studying actor-critic design choices.
Additionally, keeping the critic error bounded before each update is consistent with two-timescale convergence analysis of actor-critic methods, which require the critic to update on a faster timescale than the actor \citep{borkar1997stochastic,konda1999actor,Bhatnagar2009-NatureActorCritic}.
This method introduces two hyperparameters ($\omega$ and $n_\text{steps}$), but we find the threshold $\omega$ is easier to set than a fixed UTD ratio because it is interpretable relative to the reward scale.
$n_\text{steps}$ can be set to a reasonably high value depending on the available compute.
See Fig. \ref{fig:critic_utd} for a detailed comparison against fixed UTD ratios.
\captionsetup[algorithm]{font=small, labelfont=small, textfont=small}{
\begin{algorithm}[th]
	\small
	\caption{\small Action-value Update with Adaptive Critic UTD ratio}
	\begin{algorithmic}[1]
		\State Input: action-value parameters $w$, replay buffer $\mathcal B$
		\State Hyperparameters: error tolerance $\omega$ (default $0.001$), maximum number of critic updates $n_\text{steps}$ (default 100), batch size $|B|$
		\State Initialize error $\text{err} = \infty$, steps = 0
		\While{steps $ < n_\text{steps}$ and $\text{err} > \omega$}  
		\State steps $\gets$ steps $+ 1$
		\State Sample a mini-batch $ B$ from buffer $\mathcal B$
        \State Compute mean-squared error:
        \Statex \centerline{$\displaystyle \text{err} \;\gets\; \frac{1}{| B |} \sum_{(a,r) \in B}\bigl(r - q_w(a)\bigr)^2$}
        \State Update parameters by gradient descent:
        \Statex \centerline{$\displaystyle w \;\gets\; w - \alpha\, \nabla_{w}\, \text{err}$}
		\EndWhile
	\end{algorithmic}
	\label{alg:critic}
\end{algorithm}
}
}
\showmatmethods{} 
\dataavail{All the study data are included in this article and the SI Appendix, including the original data from the water treatment plant used to construct our simulator. 
The code is released at \url{https://github.com/haseebs/deconstruct-ac}.
}

\acknow{This work was supported by NSERC Discovery Grants (M.W., A.W.), Canada CIFAR AI Chairs Program (M.W., A.W.), and a Canada Research Chair (M.W.). 
Computational resources were provided by the Digital Research Alliance of Canada.}

\section*{References}
\bibliography{hstemplate_refs}

\onecolumn
\newpage
\beginsupplement[margin=0.95in]
    \begin{center}
        \textbf{\Huge Supplementary Materials}\\
        \vspace{1em}
    \end{center}



\begingroup
  \small
  \setlength{\parskip}{0pt}

  \let\clearpage\relax
  \let\cleardoublepage\relax
  \renewcommand{\contentsname}{Contents}
  \tableofcontents
  \vspace{-0.25em}

  \renewcommand{\listalgorithmname}{\large List of Algorithms}
  \listofalgorithms

  \renewcommand{\listfigurename}{List of Figures}
  \listoffigures
  \vspace{-0.25em}
  
  \renewcommand{\listtablename}{List of Tables}
  \listoftables
  \vspace{-0.25em}
\endgroup
\clearpage

\setcounter{figure}{0}
\renewcommand{\thefigure}{S\arabic{figure}}

\setcounter{table}{0}
\renewcommand{\thetable}{S\arabic{table}}

\setcounter{algorithm}{0}
\renewcommand{\thealgorithm}{S\arabic{algorithm}}

\section{Reward Surfaces}
\label{app:reward_surface}
\haseeb{maybe its better to remove this section heading and just let the reward surface figure float here}
\begin{figure*}[h]
		\centering
		\begin{minipage}{0.99\textwidth}
			\includegraphics[width=0.99\linewidth]{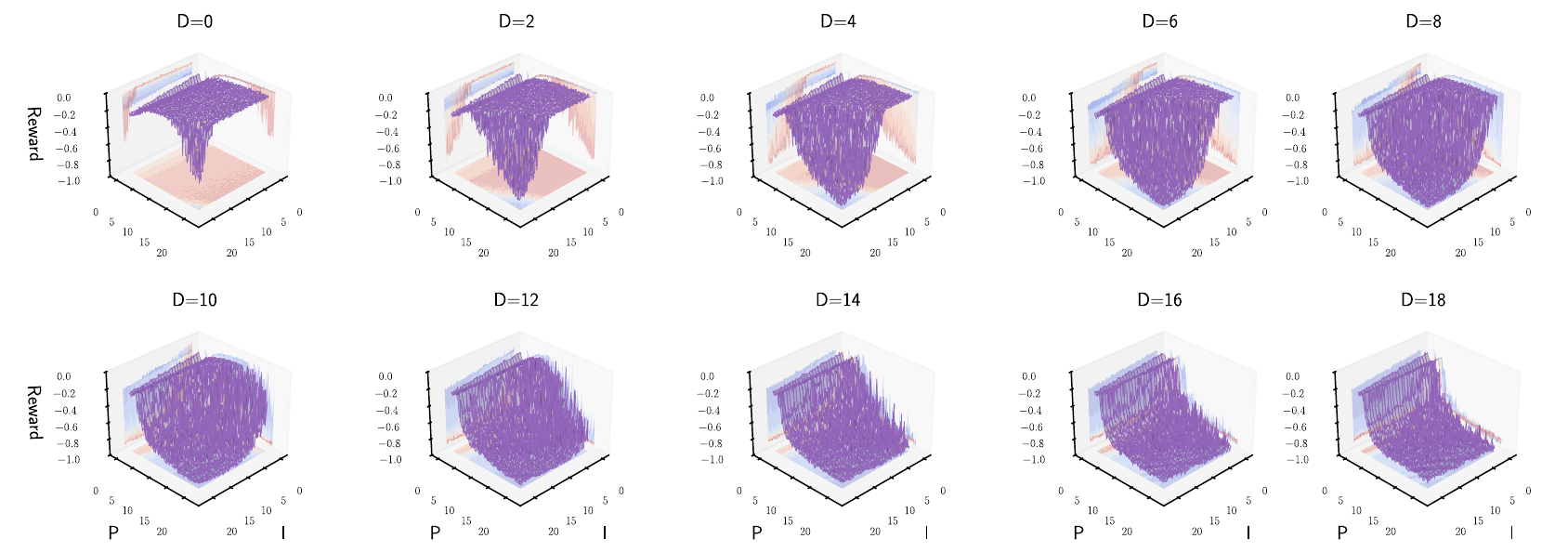}
		\end{minipage}
		\caption
		[Reward surfaces for the Backwashing-PID environment]
		{
			\textbf{The Backwashing-PID environment is non-trivial despite its simplicity.}
			The reward is the clipped negative norm of the per-step flow-rate deviation from the setpoint, with overshoots penalized more heavily than undershoots.
			Each panel fixes the derivative gain $d$ and shows the reward as a function of the proportional gain $p$ (x-axis) and integral gain $i$ (y-axis), while $d$ increases across panels.
			All three gains (actions) are bounded to $[0, 20]$.
			Three properties make this environment challenging for an actor-critic agent.
			First, most of the action space is dominated by low-reward plateaus, while the high-reward region is a narrow area that the policy must concentrate on without overshooting.
			Second, the three gains couple non-trivially: changing $d$ reshapes the entire $(p, i)$ surface, so the agent cannot learn each dimension independently.
			Third, the agent must learn under two independent sources of noise, i.e., the measurement noise $\varepsilon^{(f)}\!\sim\!\mathcal{N}(0, \sigma_f^2)$ in the flow simulator and reward noise $\varepsilon^{(r)}\!\sim\!\mathcal{N}(0, \sigma_r^2)$ in the observed return.
		}
		\label{fig:pid_surfaces}
	\end{figure*}
\clearpage

\section{Additional Conclusions}
\label{app:conclusions}


\begin{figure*}[h]
	\centering
	\includegraphics[width=0.95\linewidth]{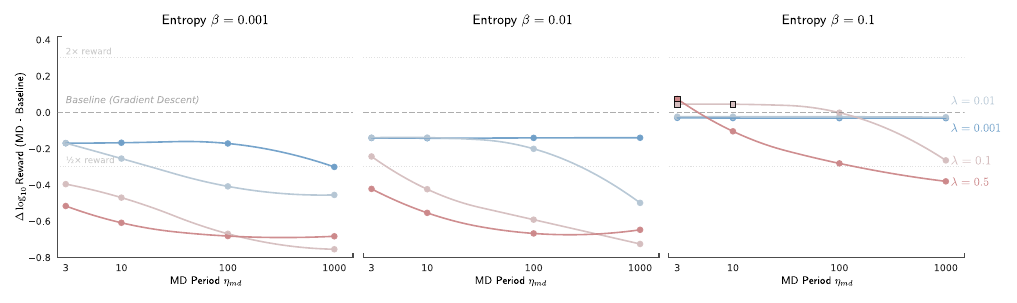}
	\caption
	[Mirror descent does not consistently improve Soft Actor-Critic]
	{\textbf{Mirror descent does not consistently improve Soft Actor-Critic. }
		The y-axis shows $\Delta \log_{10}(\text{Reward}) = \log_{10}(\text{Reward}_{\mathrm{MD}}) - \log_{10}(\text{Reward}_{\mathrm{Baseline}})$, where the rewards on the R.H.S. are the rewards averaged over the top three learning rates per hyperparameter configuration.
		The columns represent changing entropy levels $\beta$, the x-axis represents the Mirror Descent (MD) period $\eta_{md}$, and the colours represent different MD step sizes $\lambda=1/\tau$.
		The curves are drawn using a quadratic B-spline interpolation across MD periods without altering pointwise values.
		Values above the zero line represent higher average rewards under MD, while the values under represent performance worse than the Gradient Descent (GD) baseline.
		At the two lower entropies ($\beta=\{0.001, 0.01\}$), all MD configurations perform worse than GD.
		The worst case is at $1/\tau=0.5$ and high $\eta_{md}$, where the performance drops by roughly 5x compared to GD.
		At the highest entropy ($\beta=0.1$), we observe a small improvement at the lowest MD step sizes (represented by squares), but the absolute performance at this entropy is still significantly lower than the lower entropies (Fig.~\ref{fig:md_all}).}
	\label{fig:md_main}
\end{figure*}

\begin{figure*}[p]
    \centering
    \includegraphics[width=0.75\linewidth]{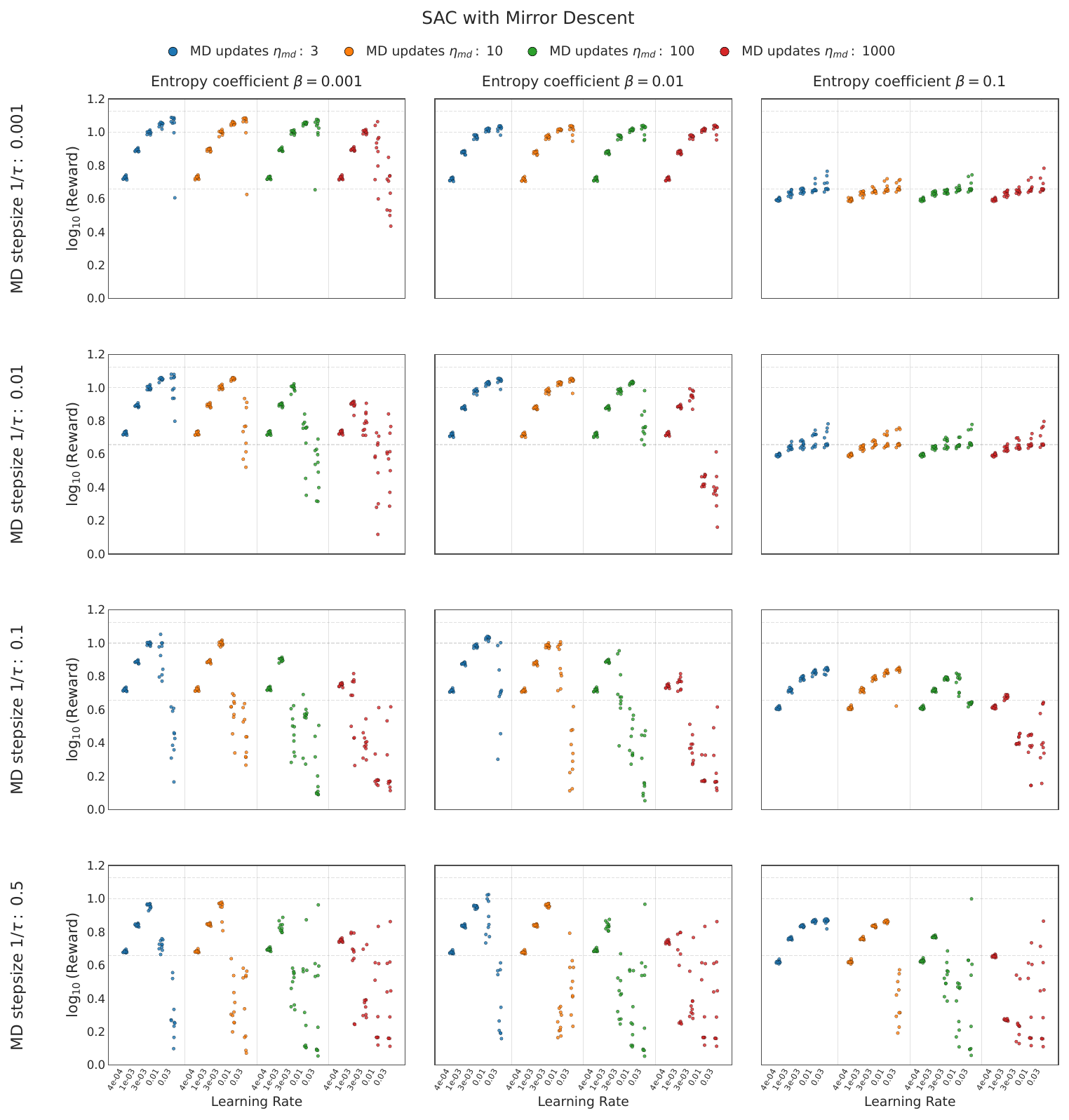}
        \caption
        [Analysis of Soft Actor-Critic with mirror descent]
        {
        \textbf{Analysis of Soft Actor-Critic with mirror descent.}
        Each dot represents the average reward of a single training run for a given learning rate.
        The columns represent changing entropy levels $\beta$, the rows represent different mirror descent stepsize $\lambda=1/\tau$, and the colours represent different mirror descent periods $\eta_{md}$.
        If we look at the lowest entropy settings $\beta= \{0.001, 0.01\}$ (first two columns), we observe that as we increase $1/\tau$ (by going down) or $\eta_{md}$ (by changing colour), the performance at every learning rate is reduced and run-to-run variability is increased.
		At the highest entropy setting $\beta = 0.1$ (rightmost column), the same trends hold but are less severe, with stronger MD settings reaching slightly higher peak rewards.
		Overall, the smallest MD step size $1/\tau=0.001$ at small $\eta_{md}$ matches the baseline SAC most closely.
		See Fig.~\ref{fig:md_main} for the summarized comparison.
        }
    \label{fig:md_all}
\end{figure*}

\begin{ourconclusion}
	Mirror descent adds complexity without consistent empirical gains
\end{ourconclusion}
\noindent 
In Fig. \ref{fig:md_main}, we observe that increasing the mirror descent period $\eta_{md}$ generally resulted in similar or significantly worse performance. 
Increasing the mirror descent step size $\lambda$ (where $\lambda=1/\tau$) resulted in worse mean rewards at smaller entropies ($\beta \in \{0.001, 0.01\}$) but higher at the largest tested entropy ($\beta=0.1$). 
However, the mean reward at the larger entropy is still significantly lower than smaller entropies (Fig. \ref{fig:md_all}).
Overall, the best performance is achieved at the smallest values of mirror descent stepsize $\lambda$ and period $\eta_{md}$.
This indicates the stronger the mirror descent, the lower is the resulting performance.

\begin{figure*}[p]
	\centering

		\includegraphics[width=0.75\linewidth]{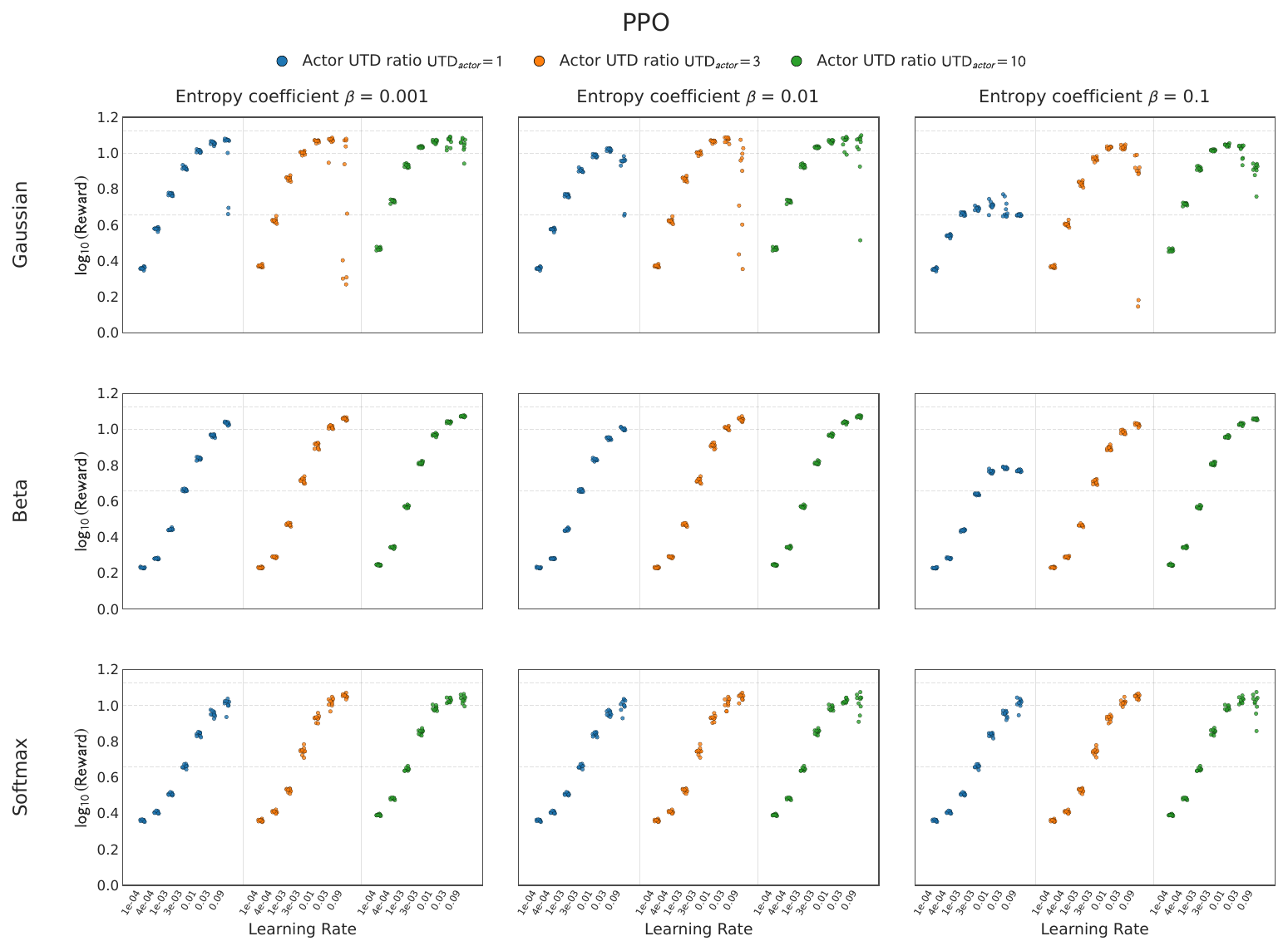}

	\caption
	[Analysis of on-policy batch size in Proximal Policy Optimization]
	{
		\textbf{Analysis of on-policy batch size in Proximal Policy Optimization.} 
		Each dot represents the average reward of a single training run for a given learning rate.
		The columns represent changing entropy levels $\beta$, the rows represent different policy parameterizations, and the colours represent different on-policy batch sizes $b$.
		The softmax policy discretizes each continuous action dimension into fixed grid of 20 bins and samples from a categorical distribution over these grid points, with per-action probabilities given by temperature-scaled softmax over learned vector of scores.
		As PPO is an on-policy algorithm, we freeze the policy for $b$ steps, collect transitions into a buffer, and then update the policy using $b$ gradient descent updates.
		We observe that both the Gaussian and beta policies suffer a performance drop as entropy regularization increases. 
		In contrast, the Softmax policy does not suffer from a performance drop.
		Across all policies, increasing the on-policy batch size $b$ improves performance, with the effect more pronounced in the high-entropy setting.
	}
	\label{fig:ppo_batch}
\end{figure*}

\begin{ourconclusion}
	Beta policy is more robust to higher learning rates when the critic's contribution is minimized
\end{ourconclusion}
\noindent 
PPO (Fig. \ref{fig:ppo_batch}) and AC($\lambda>0.9$) (Fig. \ref{fig:acl_all}) show improved robustness to higher learning rates when coupled with the beta policy.
In the bandit setting, PPO has no critic, and we observe that the beta policy at lower actor UTD ratios can perform just as well as at higher ratios when the learning rate is tuned.
In all the other settings, the beta policy struggled to learn unless the actor UTD ratio was increased.

\begin{figure*}[h!]
    \centering
    \includegraphics[width=0.99\linewidth]{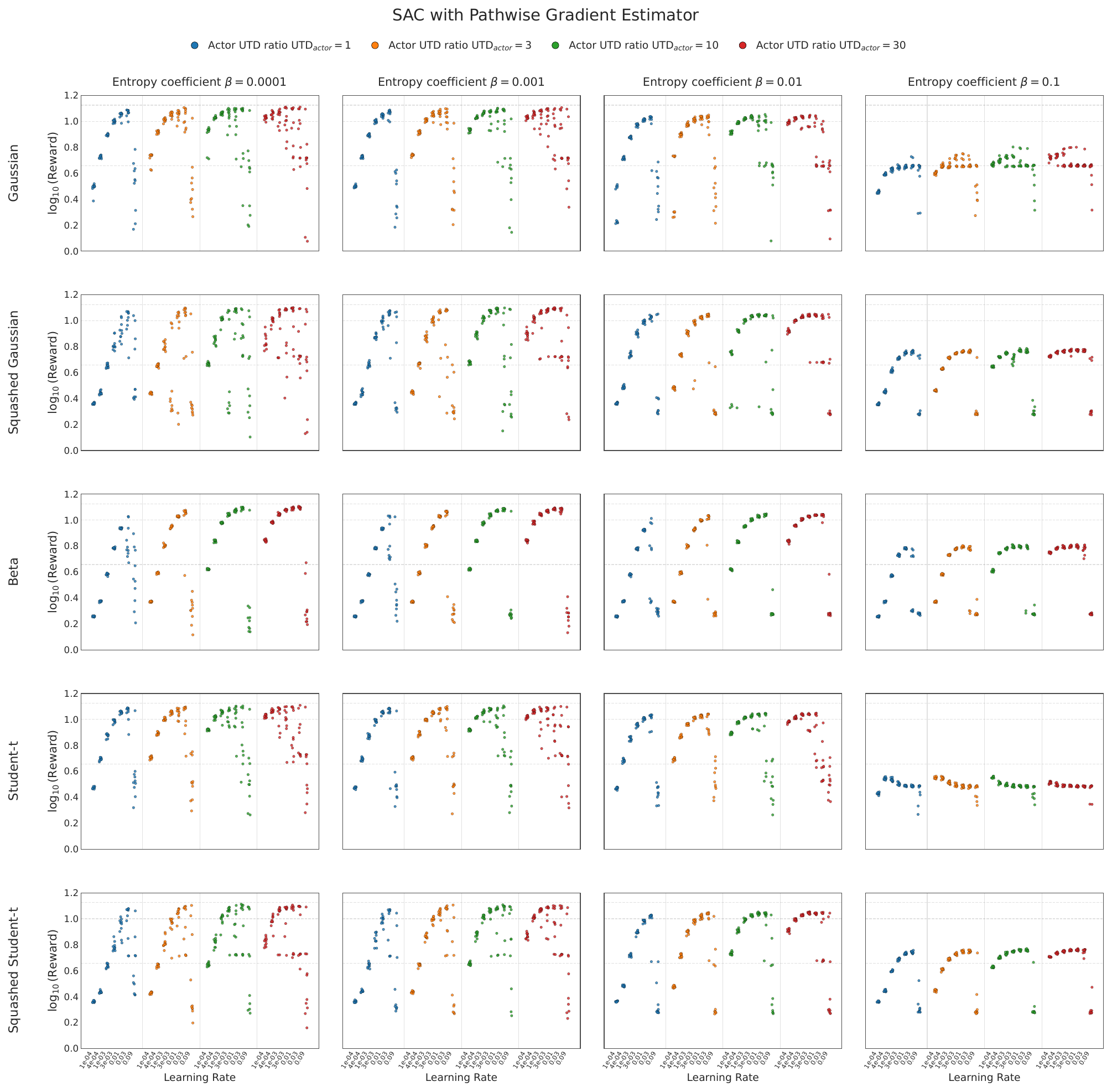}
        \caption
        [Analysis of Soft Actor-Critic with reparameterization trick (pathwise gradient)]
        {
        \textbf{Analysis of Soft Actor-Critic with reparameterization trick (pathwise gradient).}
        Each dot represents the average reward of a single training run for a given learning rate.
        The columns represent changing entropy levels $\beta$, the rows represent different policy parameterizations, and the colours represent different actor UTD ratios $\texttt{UTD}_{actor}$.
        By comparing the Gaussian (row 1) and the squashed Gaussian (row 2), we observe that the squashed Gaussian has higher run-to-run variability and is more sensitive to the learning rate, while being less sensitive to the entropy regularization hyperparameter.
        Looking at any colour and increasing entropy (going left-to-right), we observe that performance decreases significantly as entropy is increased.
		The beta policy (row 3) is the most robust, while Student's $t$ (row 4) suffers from highest degradation in performance. 
        If we analyze the actor UTD ratios (changing colours), we observe that the Gaussian, squashed Gaussian, Student's $t$ and squashed Student's $t$ policies become increasingly unstable as the ratio $\texttt{UTD}_{actor}$ increases.
        On the other hand, the beta policy performs poorly at the lowest  $\texttt{UTD}_{actor}$ and better at higher ratios.
        The highest mean performance and lowest variability are achieved with a beta policy parameterization and a high actor UTD ratio.       
        }
    \label{fig:sac_reparam}
\end{figure*}

\begin{figure*}[p]
    \centering
    \begin{minipage}{0.99\textwidth}
    \includegraphics[width=0.99\linewidth]{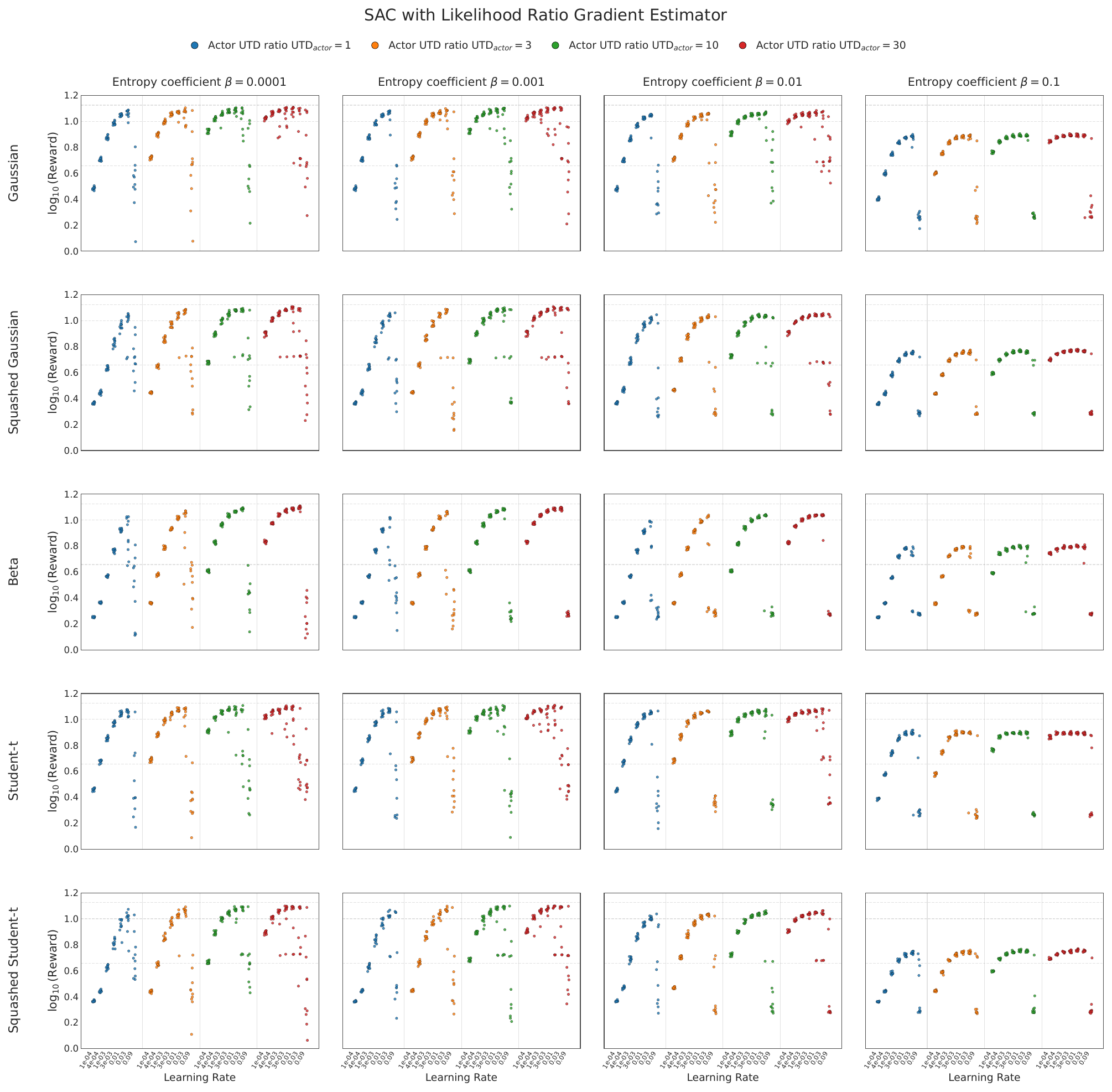}
    \end{minipage}
        \caption
        [Analysis of Soft Actor-Critic without reparameterization trick (log-likelihood gradient)]
        {
        \textbf{Analysis of Soft Actor-Critic without reparameterization trick (log-likelihood gradient).}
        Each dot represents the average reward of a single training run for a given learning rate.
        The columns represent changing entropy levels $\beta$, the rows represent different policy parameterizations, and the colours represent different actor UTD ratios $\texttt{UTD}_{actor}$.
        We observe that the Gaussian (row 1) and Student's $t$ (row 4) are the most robust to changes in the entropy coefficient hyperparameter, a behavior which is significantly different from the pathwise estimate (Fig. \ref{fig:sac_reparam}).
        If we analyze the actor UTD ratios (changing colours), we observe similar behavior: all policies other than beta (row 3) become increasingly unstable as the ratio $\texttt{UTD}_{actor}$ increases.
        In fact, we observe that the beta policy with the log-likelihood estimator can perform almost as well as the pathwise estimator shown in Fig. \ref{fig:sac_reparam} when the hyperparameters are matched, while other policies are more sensitive to the gradient estimator choice.
        }
    \label{fig:sac_noreparam}
\end{figure*}

\begin{figure*}[p]
    \centering
    \includegraphics[width=0.75\linewidth]{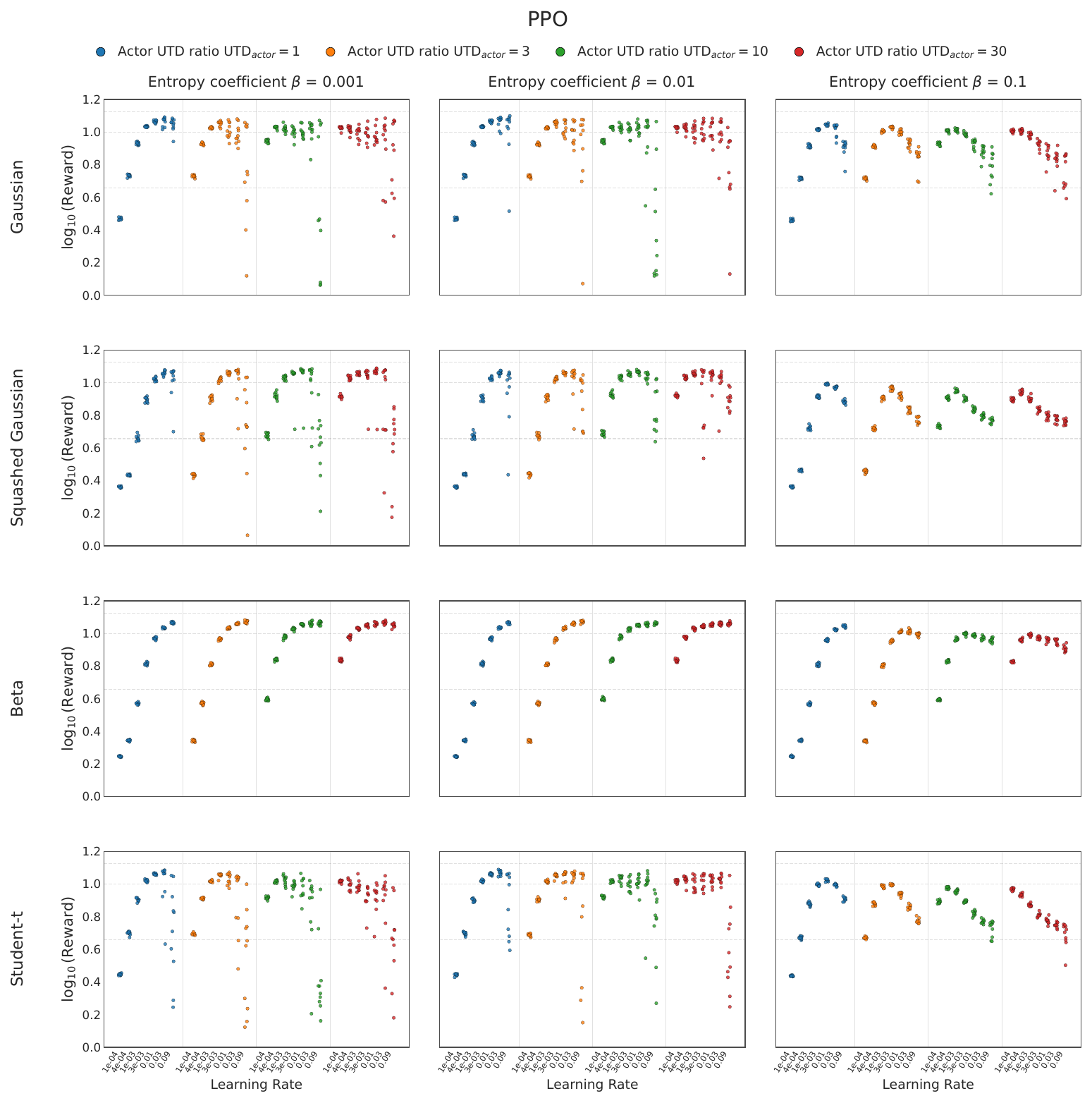}
        \caption
        [Analysis of actor UTD ratio in Proximal Policy Optimization]
        {
        \textbf{Analysis of actor UTD ratio in Proximal Policy Optimization.} 
        Each dot represents the average reward of a single training run for a given learning rate.
        The columns represent changing entropy levels $\beta$, the rows represent different policy parameterizations, and the colours represent different actor UTD ratios $\texttt{UTD}_{actor}$.
        In this experiment, the on-policy batch size is fixed at $b=10$.
        For all policies other than beta, increasing the entropy reduces the maximum achieveable performance.
        Increasing the actor UTD ratio (different colours) increases variability across runs for the Gaussian, squashed Gaussian and Student's $t$.
        In the case of beta policy, we observe that increasing $\texttt{UTD}_{actor}$ shifts the performance across all learning rates upwards, including the best learning rate.
        Additionally, we observe that the beta policy's performance drops off less significantly at the highest learning rate as compared to other policies.
        }
    \label{fig:ppo_utd}
\end{figure*}

\begin{figure*}[p]
	\centering
		\includegraphics[width=0.75\textwidth]{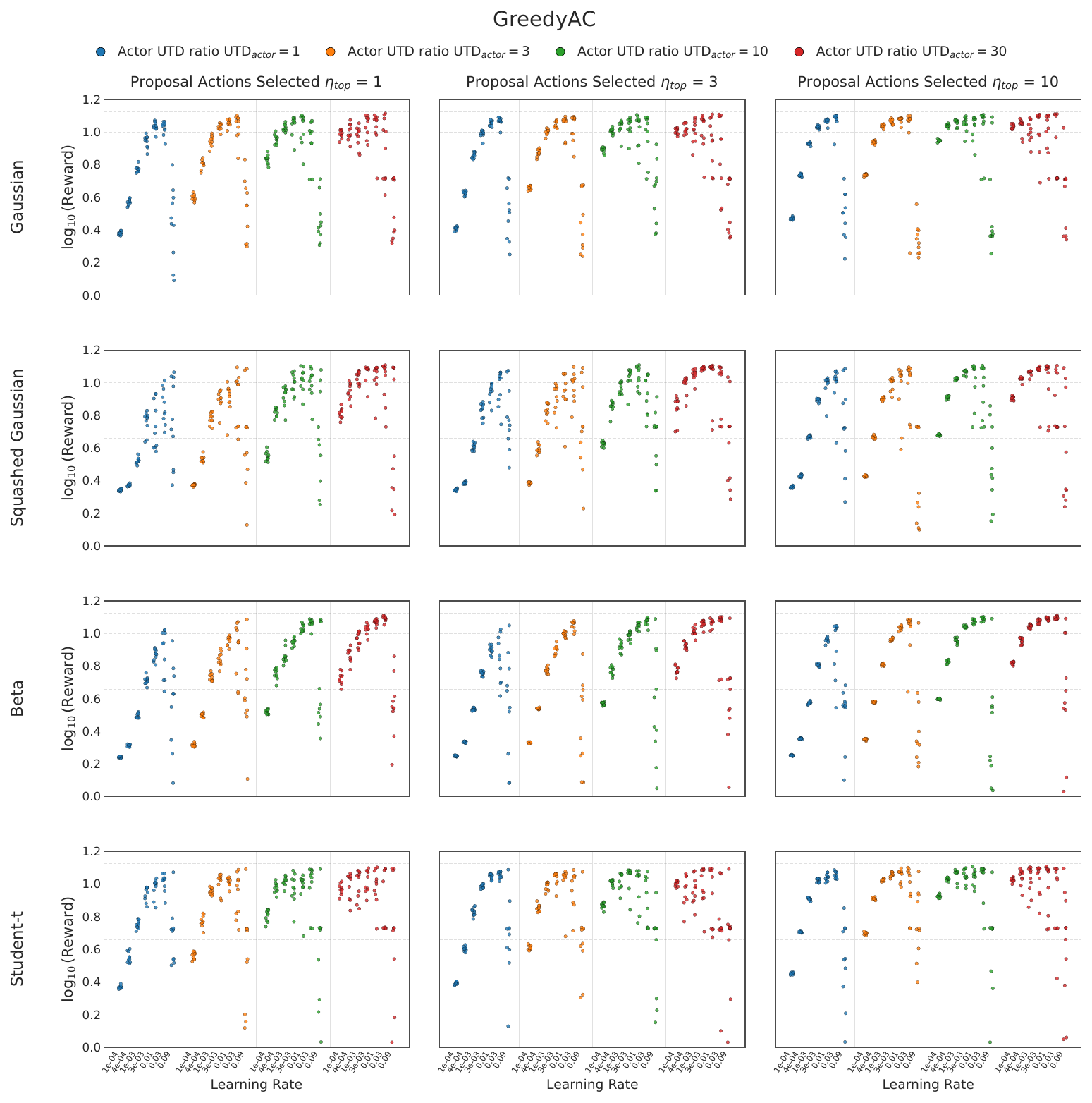}
	\caption
	[Analysis of Greedy Actor-Critic]
	{
		\textbf{Analysis of Greedy Actor-Critic.}
		Each dot represents the average reward of a single training run for a given learning rate.
		The columns represent the changing number of selected proposal actions $\texttt{UTD}_{actor}$, the rows represent different policy parameterizations, and the colours represent different actor UTD ratios $\eta_{top}$.
		Increasing $\texttt{UTD}_{actor}$ increases the run-to-run variability for all policy parameterizations, except for beta, in which case the performance improves and variability reduces.
		The Gaussian, squashed Gaussian, and Student's $t$ policies perform better at low to intermediate actor UTD ratios, whereas the beta policy performs better at a high ratio.
	}
	\label{fig:greedyac}
\end{figure*}

\begin{figure*}[p]
	\centering
	\includegraphics[width=0.99\linewidth]{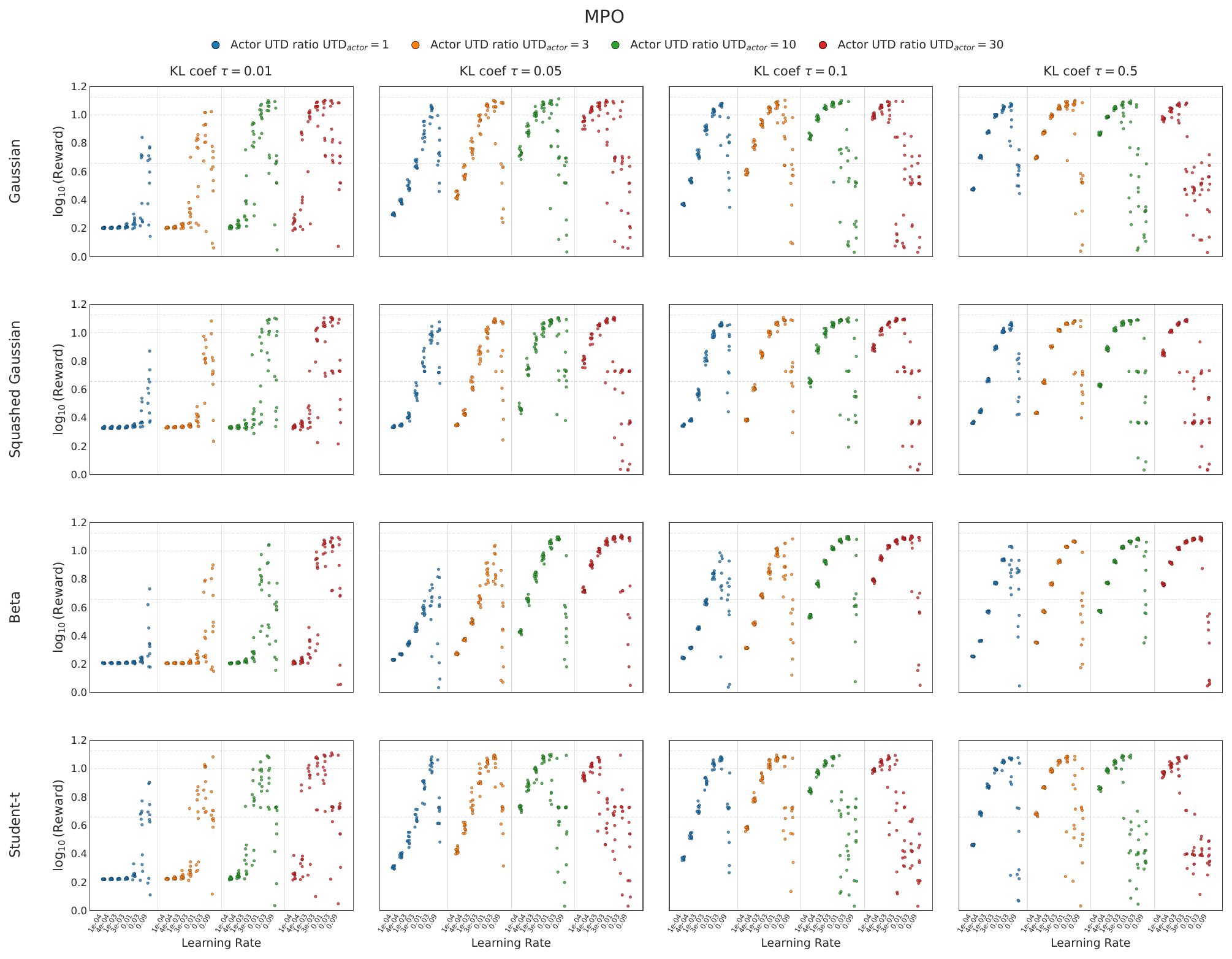}
	\caption
[Analysis of Maximum a Posteriori Policy Optimization]
{        
	\textbf{Analysis of Maximum a Posteriori Policy Optimization.}
	Each dot represents the average reward of a single training run for a given learning rate.
	The columns represent the KL coefficient $\tau$, the rows represent the different policy parameterizations, and the colours represent different actor UTD ratios $\texttt{UTD}_{actor}$.
	For all policies, only a narrow range of hyperparameters perform well.
	If we analyze the actor UTD ratios (changing colours), we observe that the Gaussian, squashed Gaussian, Student's $t$ and squashed Student's $t$ policies become increasingly unstable as the ratio $\texttt{UTD}_{actor}$ increases.
	On the other hand, the beta policy performs poorly at the lowest  $\texttt{UTD}_{actor}$ and better at higher ratios.
}
\label{fig:mpo}
\end{figure*}

\begin{ourconclusion}
	Actor UTD reshapes learning-rate sensitivity
\end{ourconclusion}
\noindent 
The relationship between learning rate and mean performance is typically similar to an upside-down parabola: with the best setting in the middle, and a poorer performance at the higher and lower extremes.
We observe that increasing the actor UTD ratio can alter the shape of this parabola, and in some cases, reduce the sensitivity to learning rate hyperparameter.
This implies that in certain settings, an arbitrarily chosen learning rate from a reasonable range is more likely to perform well.
We observe this reduction in sensitivity with all the policy losses that we used to analyze $\texttt{UTD}_{actor}$: SAC with both gradient estimators (Figs. \ref{fig:sac_reparam}-\ref{fig:sac_noreparam}), PPO (Fig. \ref{fig:ppo_utd}), GreedyAC (Fig. \ref{fig:greedyac}) and MPO (Fig. \ref{fig:mpo}).
However, the run-to-run variability depends on the policy parameterization.
As $\eta_{actor}$ is increased, the Gaussian, Student's $t$, and squashed policies generally exhibit higher variabilities, while the variability of the beta policy is reduced.
Due to this, the reduced learning-rate sensitivity is useful only with the beta policy parameterization.

\begin{ourconclusion}
	Averaging over more candidate actions stabilizes policy updates
\end{ourconclusion}
\noindent Across multiple policy losses, we observe that increasing the number of samples used to compute the policy gradient increases the performance and/or reduces overall run-to-run variability.
In PPO, increasing the on-policy batch size increases mean performance, particularly at higher entropies (Fig. \ref{fig:ppo_batch}).
We do not observe an improvement in variability for PPO because it is already too small.
In GreedyAC, increasing the number of proposal actions $\eta_{top}$ both improves the performance and reduces variability (Fig. \ref{fig:greedyac}).
Although the improvements are monotonic within our tested ranges, we surmise that there can be diminishing returns or adverse effects if these hyperparameters are increased further.



\clearpage

\section{Policy Parameterizations}
\label{app:params}

 \textbf{Gaussian Policy:} One of the most common policy parameterizations in the literature is based on the Gaussian distribution, which assumes that the ideal actions are normally distributed around a learned mean. In the scalar bandit setting, a policy using this reparameterization can be written as:
  \begin{equation}
    \pi_\pparams(a) \doteq \frac{1}{\sigma_\pparams\sqrt{2\pi}} \exp\left( -\frac{(a - \mu_\pparams)^2}{2\sigma_\pparams^2} \right)
    \label{eq:gaussian}
  \end{equation}
  where $\mu_\pparams \in \mathbb{R}$ is the mean and $\sigma_\pparams \in \mathbb{R}^{+}$ is the standard deviation. To ensure positivity, the standard deviation is typically parameterized as an exponential of an unconstrained parameter, e.g., $\sigma_\pparams = \exp(\sigma_{\text{raw}})$. The Gaussian distribution has infinite support, which is not practical for policy gradient algorithms. We will discuss two methods of restricting the support and sampling from a Gaussian:

  \textit{1) Clipped Gaussian Sampling.} In this approach, the raw mean $\mu_{\text{raw}}$ is passed through a $\tanh$ nonlinearity and scaled to fit within the action bounds:
\begin{equation} \label{eq:gaussian_sample1}
  \mu = a_{\min} + \frac{a_{\max} - a_{\min}}{2} \left( \tanh(\mu_{\text{raw}}) + 1 \right)
\end{equation}

\begin{equation} \label{eq:gaussian_sample2}
  a = \mu + \sigma \cdot \epsilon, \quad \epsilon \sim \mathcal{N}(0, 1)
\end{equation}

\begin{equation} \label{eq:gaussian_sample3}
  a_{\text{scaled}} = \text{clip}(a, a_{\min}, a_{\max})
\end{equation}


  \textit{2) Squashed Gaussian Sampling.} An alternative approach is to sample from the Gaussian and apply a smooth $\tanh$ squashing function:
 
\begin{equation}
  \tilde{a} = \mu + \sigma \cdot \epsilon, 
  \quad \epsilon \sim \mathcal{N}(0, 1), 
  \quad a = \tanh(\tilde{a})
\end{equation}
followed by rescaling:
\begin{equation}
  a_{\text{scaled}} = a_{\min} + \frac{a_{\max} - a_{\min}}{2} \left( a + 1 \right)
\end{equation}

The squashed Gaussian applies tanh directly to the sampled action, which alters the distribution's support and requires a log-determinant Jacobian correction to maintain accurate log-probability estimates \citep{haarnoja-SAC2018}.
\\
\\
\textbf{Beta Policy}
The beta distribution belongs to non-location-scale family of distributions and has a restricted support in the interval $[0,1]$, which ensures that no probability density falls outside the boundary and eliminates the need for any clipping or squashing. This distribution offers flexibility in modelling peaked, skewed, or bi-modal distributions within a bounded domain. A policy using the beta parameterization can be defined as:
\begin{equation}
\pi_\pparams(a) \doteq \frac{a^{\alpha_\pparams - 1}(1 - a)^{\beta_\pparams - 1}}{B(\alpha_\pparams, \beta_\pparams)}, \quad a \in [0, 1]
\end{equation}

where $\alpha_\pparams, \beta_\pparams \in \mathbb{R}^{+}$ are shape parameters, and $B(\alpha_\pparams, \beta_\pparams)$ is the beta function defined as:
\begin{equation}
B(\alpha_\pparams, \beta_\pparams) = \frac{\Gamma(\alpha_\pparams)\Gamma(\beta_\pparams)}{\Gamma(\alpha_\pparams + \beta_\pparams)}
\end{equation}

where $\Gamma(\cdot)$ is the gamma function. In a beta policy, the actions are sampled and scaled as:
\begin{equation}
\begin{aligned}
  a &\sim B(\alpha_\pparams, \beta_\pparams) \\
  a_{\text{scaled}} &= a_{\min} + (a_{\max} - a_{\min}) \cdot a
\end{aligned}
\end{equation}

Similar to the Gaussian policy, the beta policy also supports the reparameterization trick. In this work, we restrict $\alpha_\pparams>1; \beta_\pparams>1$, which ensures that the policy is unimodal.
\\
\\
\textbf{Student's $t$ Policy}
The Student's $t$ distribution allows for heavier tails and is more robust to outliers than the Gaussian distribution. A policy using the Student's $t$ parameterization is given by:
\begin{equation}
\pi_\pparams(a) \doteq \frac{\Gamma\left(\frac{\nu_\pparams + 1}{2}\right)}{\sqrt{\nu_\pparams\pi}\sigma_\pparams \Gamma\left(\frac{\nu_\pparams}{2}\right)} \left( 1 + \frac{(a - \mu_\pparams)^2}{\nu_\pparams\sigma_\pparams^2} \right)^{-\frac{\nu_\pparams + 1}{2}}
\end{equation}
where $\mu_\pparams \in \mathbb{R}$ is the location, $\sigma_\pparams \in \mathbb{R}^{+}$ is the scale, and $\nu_\pparams \in \mathbb{R}^+$ is the degrees of freedom. The mean exists only when $\nu_\pparams > 1$, and the variance exists only when $\nu_\pparams > 2$. The sampling and bounding of support is done similarly to the clipped Gaussian, as described by Eq. \ref{eq:gaussian_sample1},\ref{eq:gaussian_sample2},\ref{eq:gaussian_sample3}.

Compared to the Gaussian policy, the Student's $t$ policy produces heavier-tailed action distributions, which can promote better exploration-exploitation \citep{Zhu2025-qExpPolicy,Zhu2025-FatToThin}. 
An important special case is $\nu_\pparams = 1$, which corresponds to the Cauchy distribution, an extremely heavy-tailed distribution without a defined mean or variance. As $\nu_\pparams \to \infty$, the Student's $t$ distribution converges to a relatively lighter-tailed Gaussian, recovering the standard Gaussian policy as a limiting case. In practice, a Student's $t$ policy with $\nu_\pparams \geq 30$ is sufficiently similar to the Gaussian policy.
\\
\\
\textbf{Deterministic Policy}
A deterministic policy directly outputs an action:
\begin{equation}
a = \mu_\pparams.
\end{equation}
This approach is used in deterministic actor-critic methods such as DDPG. Although it eliminates action stochasticity, external noise (e.g., Ornstein-Uhlenbeck or Gaussian noise) is typically added during training to encourage exploration.

\clearpage
\section{Actor-Critic Algorithms}
\label{sec:loss}


\subsection*{Critic Update}
\label{appendix:critic_section}
All of the algorithms that we discuss share the same adaptive critic update as described in the main text. 
In practice, however, most implementations utilize a fixed critic UTD ratio $\texttt{UTD}_{critic}$, which involves repeatedly sampling from a replay buffer and updating the critic network using regression for $\texttt{UTD}_{critic}$ steps.
This procedure is described in Alg. \ref{alg:critic_fixed_utd}.

\begin{algorithm}[th]
	\caption{Action-value update with fixed critic UTD ratio}
	\begin{algorithmic}[1]
		\State Input: Action-value parameters $w$, replay buffer $\mathcal B$
		\State Hyperparameters: critic UTD ratio $\texttt{UTD}_{critic}$, batch size $|B|$
		\For{steps $=1,2,3, \dots, \texttt{UTD}_{critic}$}
		\State Sample a mini-batch $ B$ from buffer $\mathcal B$
        \State Compute mean-squared error:
        \Statex \centerline{$\displaystyle \text{err} \;\gets\; \frac{1}{| B |} \sum_{(a,r) \in B}\bigl(r - q_w(a)\bigr)^2$}
        \State Update parameters by gradient descent:
        \Statex \centerline{$\displaystyle w \;\gets\; w - \alpha\, \nabla_{w}\, \text{err}$}
		\EndFor
	\end{algorithmic}
	\label{alg:critic_fixed_utd}
\end{algorithm}

\subsection*{Actor Critic $\lambda$}
On-policy AC($\lambda)$ allows us to range from an unbiased, higher variance REINFORCE update, with $\lambda = 1$, to a more biased, lower-variance actor-critic algorithm with $\lambda = 0$ that only uses $q_\qparams$ with no reward sample in the actor update. The algorithm is naturally on-policy, because we update with the reward from the action that the agent took. If we wanted to update from $(a,r)$ stored in the replay buffer, then we would have to use importance sampling to correct for the action distribution. It is rare to do so, and instead, to use $\lambda$-returns, it is typically more common to simply run on-policy algorithms like PPO. Note that on-policy AC($\lambda$) is like PPO in this way, but does not have all the extra tricks related to clipping. The complete algorithm is described in Alg. \ref{alg:acl}.

\begin{algorithm}[h!]
\caption{Actor update for  AC($\lambda)$ }
\begin{algorithmic}[1]
    \State Initialize: policy parameters $\theta$, action-value parameters $w$, empty replay buffer $\mathcal B$, value baseline $v = 0$
    \State Hyperparameters: Value-reward mixing coefficient $\lambda \in [0,1]$, value baseline decay rate $\eta_v$ (default $0.1$), actor and critic stepsizes

    \For{$t=1,2,3 \dots$}
   		\State Sample action $A_t \sim \pi_\theta$
		\State Execute action $A_t$, observe reward $R_t$
        \State Update value baseline: $v \gets (1-\eta_v) v + \eta_v R_{t} $
	\State Store $ (A_t,R_t)$ in replay buffer $\mathcal B$
        \State Update $q_\qparams$ using Algorithm \ref{alg:critic_fixed_utd}
        \State Update $\pi_\pparams$ by gradient ascent using gradient  $\delta \nabla_\pparams \ln \pi_\pparams(a)$ for $\delta \doteq (\lambda R_t + (1-\lambda) q_\qparams(A_t)) - v$
    \EndFor
\end{algorithmic}
\label{alg:acl}
\end{algorithm}

\clearpage
\subsection*{Deep Deterministic Policy Gradient}
DDPG learns a deterministic policy. For a bandit, this means it's policy parameters $\pparams$ correspond to an estimate of the best action. DDPG is an off-policy method that learns the greedy policy ($\pparams$), and adds stochastic noise for exploration. In particular, we use Gaussian noise, with clipping to ensure the action remains in the bounded range. For consistency in the pseudo-code, we still write $A_t \sim \pi_\pparams$, which in this case means we sample $\epsilon_t \sim \mathcal{N}(0, \sigma)$ for noise variance $\sigma > 0$, and set $A_t = \text{clip}_{\mathcal{A}}(\pparams + \epsilon_t)$. 

The update for DDPG modifies the policy parameters by taking the gradient $\nabla_\pparams q_\qparams(\pi_\pparams)$, to adjust the policy parameters to get a higher value on $q_\qparams$ (to get an action where $q_\qparams(a)$ is higher). Using the chain rule, we take $\nabla_a q_\qparams(a)$ and then look at how the action changes as $\pparams$ changes. In the bandit setting, $\pparams$ actually equals the action, so the update for the current parameters $\pparams_t$ uses $\nabla_\pparams q_\qparams(\pparams_t)$ which is the same as $\nabla_a q_\qparams(\pparams_t)$.

\begin{algorithm}[h!]
\caption{Actor update for DDPG}
\begin{algorithmic}[1]
    \State Initialize: policy parameters $\theta$, action-value parameters $w$, value baseline $v=0$
    \State Hyperparameters: Gaussian noise variance $\sigma$, Gaussian noise decay rate $\zeta$, batch size $b$
    \For{$t=1,2,3 \dots$}
    \State Sample $A_t = \text{clip}_{\mathcal{A}}(\pparams + \epsilon_t)$ with $\epsilon_t \sim \mathcal{N}(0, \sigma)$
    \State Execute action $A_t$, observe reward $R_t$
    \State Store $ (A_t,R_t)$ in replay buffer $\mathcal B$
                \State Update $q_\qparams$ using Algorithm \ref{alg:critic_fixed_utd}
    \State Update $\pparams$ with gradient ascent step using gradient $\nabla_\theta q_\qparams(\pparams)$
    \State Anneal noise using $\sigma = \sigma * \zeta$
    \EndFor

\end{algorithmic}
\label{alg:ddpg}
\end{algorithm}

\subsection*{Proximal Policy Optimization}
At each iteration $t$, PPO performs multiple steps of SGD on the objective function of the following unconstrained optimization problem:
\begin{equation*}
\max_{\pi} \,\, \mathbb{E}_{ a \sim \pi_{\theta_t}} \AdaRectBracket{ 
\min\!\AdaBracket{\frac{\pi_\theta(a)}{\pi_{\theta_t}(a)} (R-v), \,\, \text{clip}\!\AdaBracket{\frac{\pi_\theta(a)}{\pi_{\theta_t}(a)},  1-\epsilon, 1+\epsilon} (R-v)}}.
\end{equation*}
Recall that for our bandit setting there is no state and therefore the advantage function degenerates to one-sample Monte Carlo estimate $(R_t - v)$, where $v$ is a scalar baseline maintained as an exponential moving average of observed rewards.
The hyper-parameter $ \epsilon $ determines how the policy ratio $ \pi_\theta / \pi_{\theta_t} $ is clipped. 
Compared to other actor-critic algorithms, the clipping stands out as it allows the trust region to be always satisfied.

\begin{algorithm}[h!]
\caption{Actor update for PPO}
\begin{algorithmic}[1]
    \State Initialize: policy parameters $\theta$, value baseline $v=0$, empty on-policy replay buffer $\mathcal B$
    \State Hyperparameters: number of actor updates $\texttt{UTD}_{actor}$ (default 1), entropy coefficient $\beta$, clipping threshold $\epsilon$, on-policy minibatch size $b$, baseline learning rate $\eta_v$, entropy sample count $M$
    \For{$t=1,2,3 \dots$}
    \State Sample action $A_t \sim \pi_\theta$
    \State Execute action $A_t$, observe reward $R_t$
    \State Update baseline $v \leftarrow (1-\eta_v) v + \eta_v R_t$
    \State Store $(A_t, R_t, \ln \pi_{\bar\theta}(A_t))$ in on-policy replay buffer $\mathcal B$
    
    \If{\text{mod}$(|\mathcal{B}|, b) == 0$}
    \For{steps $=1,2,3, \dots, \texttt{UTD}_{actor} \cdot b$}
        \State Get samples $\{a_i\}_{1:b}, \,\, \{r_i\}_{1:b}, \,\, \{\ln \pi_{\bar\theta}(a_i)\}_{1:b}$
        \State Normalize rewards $r_i \leftarrow (r_i - \bar r)/(\sigma_r + 10^{-7})$
        \State Compute advantage $\hat\delta_i = r_i - v$
        \State Evaluate new log-probabilities: $\{\ln \pi_{\theta}(a_i)\}_{1:b}$
        \State Compute ratio $\rho_i = \exp\!\left(\ln\pi_{\theta}(a_i) - \ln \pi_{\bar\theta}(a_i)\right)$
        \State Policy loss $\mathcal{L}_{\theta} = -\frac{1}{b} \sum_{i} \min\!\left\{ \rho_i \hat\delta_i, \,\, \text{Clip}\{\rho_i, 1-\epsilon, 1+\epsilon\} \hat\delta_i \right\}$
        \State Estimate entropy $\mathcal{H}(\pi_{\theta}) \approx -\frac{1}{M}\sum_{j=1}^{M} \ln \pi_\theta(\tilde a_j)$ with $\tilde a_j \sim \pi_\theta$
        \State Update $\theta$ using gradient descent on $\mathcal{L}_{\theta} - \beta \cdot \mathcal{H}(\pi_{\theta})$
    \EndFor
        \State Empty on-policy replay buffer $\mathcal B$
    \EndIf
    \EndFor
\end{algorithmic}
\label{alg:ppo}
\end{algorithm}
\clearpage

\subsection*{Soft Actor-Critic}

As introduced before, SAC  \citep{haarnoja-SAC2018} encourages exploration by maximizing the Shannon entropy augmented objective that leads to the Boltzmann policy.
This leads to minimizing a reverse KL loss
\begin{align*}
\mathcal{L}_\text{SAC}(\theta) :&= \KL{\pi_{\theta}}{{\pi}_{\text{ent}}} \\
&= \KL{\pi_{\theta}}{\frac{\exp\AdaBracket{{q_w}(a)/{\beta}}}{Z}} \\
&= \expectation{\pi_\theta}{\ln\pi_\theta - \frac{q_w(a) - v}{\beta}} ,
\end{align*}
The parametrized policy $\pi_{\theta}$ is projected to be close to the Boltzmann policy.
In practice, $\pi_{\theta}$ is chosen to be the Gaussian policy by default, but potentially a more exploring policy like the Student's $t$ could lead to better performance.
Depending on action values, $\pi_\text{ent}$ can have multiple modes and heavy tails.
The Gaussian may not be able to fully capture these characteristics.

A mirror descent version of SAC penalizes going too far from the previous policy $\pi_t$ by adding a KL penalty $\text{KL}(\pi_\theta || \pi_t)$.
Note that the reference policy here $\pi_t$ is different from $\pi_\text{ent}$:
\begin{align*}
\mathcal{L}_\text{SAC-MD}(\theta) :&= \expectation{\pi_\theta}{\ln\pi_\theta - \frac{q_w(a) - v}{\beta} - \tau \ln\frac{\pi_\theta(a)}{\pi_t(a)}} ,
\end{align*}
where $\tau$ denotes the weighing coefficient.

\begin{algorithm}[h!]
\caption{Actor update for Soft Actor-Critic }
\begin{algorithmic}[1]
    \State Initialize: policy parameters $\pparams$, action-value parameters $\qparams$, empty replay buffer $\mathcal B$, value baseline $v = 0$
    \State Hyperparameters: number of actor updates $\texttt{UTD}_{actor}$ (default 1), actor and critic stepsizes, number actions to sample $m$, entropy coefficient $\beta$, value baseline decay rate $\eta_v$ (default $0.1$), policy from the last update (for mirror descent) $\pi_t$

    \For{$t=1,2,3 \dots$}
   	\State Sample action $A_t \sim \pi_\theta$
	\State Execute action $A_t$, observe reward $R_t$
         \State Update value baseline: $v \gets (1-\eta_v) v + \eta_v R_{t} $
	\State Store $ (A_t,R_t)$ in replay buffer $\mathcal B$
        \State Update $q_\qparams$ using Algorithm \ref{alg:critic_fixed_utd}
        \For{steps $=1,2,3, \dots, \texttt{UTD}_{actor}$}
            \State Sample $m$ actions from policy $a_1 \dots a_{m} \sim \pi_{\theta}$
            \State Update policy by gradient ascent using: 
            \Statex \vspace{1em}\centerline{$\displaystyle \frac{1}{m}\sum_{a\in a_1 \dots a_m} \big( (q_w(a) - v) - \beta \ln \pi_{\pparams}(a) \big) \nabla_\pparams \ln \pi_\pparams(a)$}
            \State or Mirror Descent:
            \Statex \vspace{1em}\centerline{$\displaystyle \frac{1}{m}\sum_{a\in a_1 \dots a_m} \left( (q_w(a) - v) - \beta \ln \pi_{\pparams}(a) - \tau \ln\frac{\pi_\theta(a)}{\pi_t(a)}\right) \nabla_\pparams \ln \pi_\pparams(a) $}
            \State $\pi_t \gets \pi_\theta$
        \EndFor
    \EndFor
\end{algorithmic}
\label{alg:sac}
\end{algorithm}

\clearpage
\subsection*{Maximum a Posteriori Policy Optimization}

MPO is also derived from the KL regularization perspective:
\begin{equation*}
\pi_{\text{kl}} \doteq \arg\max_{\pi} \mathbb{E}_\pi[q(A)] - \alpha \text{KL}(\pi || \pi_t)
\end{equation*}
where the reference policy $\pi_t$ denotes the last policy distribution.
The solution is $\pi_\text{kl}(a) \propto \pi_t(a)\exp\AdaBracket{q_w(a)/\tau}$.
MPO leverages this fact in minimizing the KL loss between $\pi_\theta$ and $\pi_\text{kl}$:
\begin{align*}
&\min_{\pi_\theta} \text{KL}(\pi_\text{kl} || \pi_\theta)\\
&=\min_{\pi_\theta} \expectation{a\sim\pi_\text{kl}}{\ln\pi_\text{kl}(a) - \ln\pi_\theta(a)}\\
&=\min_{\pi_\theta} \expectation{a\sim\pi_t}{\exp\AdaBracket{\frac{q_w(a)}{\tau}} \AdaBracket{\ln\pi_\text{kl}(a) - \ln\pi_\theta(a)}}\\
&= \min_{\pi_\theta} \expectation{a\sim\pi_t}{-\exp\AdaBracket{\frac{q_w(a)}{\tau}} \ln\pi_\theta(a)}
\end{align*}
where the last equation is because the term $\ln\pi_\text{kl}$ does not depend on the optimization variable $\theta$.
Therefore, the MPO actor objective can be understood as maximizing the log-likelihood weighted by the exponential of action value.
The original MPO algorithm adds another KL penalty to the loss above and makes the coefficient $\alpha$  learnable:
\begin{align*}
    \min_{\alpha>0} \, \max_{\pi_\theta} \expectation{a\sim\pi_t}{\exp\AdaBracket{\frac{q_w(a)}{\tau}} \ln\pi_\theta(a)} + \alpha \AdaBracket{\epsilon - {\text{KL}\AdaBracket{\pi_t || \pi_\theta}}},
\end{align*}
where $\epsilon$ denotes the trust region threshold, $\alpha$ denotes the weighing coefficient.
The complete algorithm is described in Alg. \ref{alg:mpo}.

\begin{algorithm}[h!]
	\caption{Actor update for MPO}
	\begin{algorithmic}[1]
		\State Initialize: policy parameters $\theta$, action-value parameters $w$, value baseline $v$, policy from last update $\pi_t$
		\State Hyperparameters: number of actor updates $\texttt{UTD}_{actor}$ (default 1), number of action proposals $m$, KL coefficient $\tau$, trust region threshold $\epsilon$, dual variable learning rate $\eta_\alpha$, value baseline decay rate $\eta_v$ (default $0.1$)
		    \For{$t=1,2,3 \dots$}
		\State Sample action $A_t \sim \pi_\theta$
		\State Execute action $A_t$, observe reward $R_t$
		\State Update value baseline: $v \gets (1-\eta_v) v + \eta_v R_{t} $
		\State Store $ (A_t,R_t)$ in replay buffer $\mathcal B$
		\State Update $q_\qparams$ using Algorithm \ref{alg:critic_fixed_utd}
		\For{steps $=1,2,3, \dots, \texttt{UTD}_{actor}$}
		\State Sample $m$ actions from last policy $a_1 \dots a_{m} \sim \pi_{t}$
		\State Compute scale $\exp\left(\frac{q_{w}(\{a\}_{1:m}) - v}{\tau}\right)$
    	\State Update policy by minimizing the loss: 
		\Statex \vspace{1em}\centerline{$\displaystyle \frac{1}{m} \sum_{a\in a_1 \dots a_m} \left(-\exp\left(\frac{q_{w}(a) - v}{\tau}\right)\ln\pi_{\theta}(a) - \alpha\left(\epsilon - \ln\frac{\pi_t(a)}{\pi_\theta(a)}\right)\right)$}
		\State Update coefficient $\alpha$ by minimizing the following loss using learning rate $\eta_\alpha$
		\Statex \vspace{1em}\centerline{$\displaystyle \nabla_{\alpha} \frac{1}{m} \sum_{a\in a_1 \dots a_m} \alpha\left( \epsilon - \ln\frac{\pi_t(a)}{\pi_\theta(a)}\right)$}
		\State Clamp $\alpha \gets \max(\alpha, 0)$
		\State $\pi_t \gets \pi_\theta$
		\EndFor
		\EndFor
	\end{algorithmic}
	\label{alg:mpo}
\end{algorithm}

\clearpage

\subsection*{Greedy Actor-Critic}
GreedyAC \citep{Neumann2023-greedyAC} aims to learn from unbiased rewards but at the same time enjoys expedited exploration from Shannon entropy augmented rewards.
To this end, GreedyAC maintains an additional proposal policy for exploration by maximizing Shannon entropy augmented rewards. 
Its actor policy maximizes unbiased reward and learns from the high-quality actions generated by the proposal policy. 
To simplify notations, we use $I$ to denote the set of high quality actions.
\begin{align*}
    &\mathcal{L}_\text{GreedyAC, prop}(\theta'):= \expectation{a\in I}{-\ln \pi_{\theta'}(a) - \entropyany{\pi_{\theta'}} }, \\
    &\mathcal{L}_\text{GreedyAC, actor}(\theta):= \expectation{a\in I}{-\ln \pi_\theta(a) } .
\end{align*}
GreedyAC maximizes log-likelihood of the actor and entropy-augmented likelihood for the proposal policy. 
Note that when $\pi_{\theta}$ is a non-standard policy like the beta or Student's $t$, its Shannon entropy $\entropyany{\pi_{\theta}}$ may not have a closed-form expression.
Therefore, we can use log-probabilities as a surrogate just like in SAC.
\newcommand{\ntop}{n_{\text{top}}}
\begin{algorithm}[h!]
\caption{Actor update for Greedy Actor-Critic}
\begin{algorithmic}[1]
    \State Initialize: policy parameters $\pparams$, proposal policy parameters $\pparams'$, action-value parameters $\qparams$, empty replay buffer $\mathcal B$
    \State Hyperparameters: number of actor updates $\texttt{UTD}_{actor}$ (default 1), actor and critic stepsizes, number actions to sample $m$

    \State // Use top 0.1 percentile of actions
    \State $\ntop \doteq \lceil 0.1 m \rceil$
    \For{$t=1,2,3 \dots$}
   	\State Sample action $A_t \sim \pi_\theta$
	\State Execute action $A_t$, observe reward $R_t$
	\State Store $ (A_t,R_t)$ in replay buffer $\mathcal B$
        \State Update $q_\qparams$ using Algorithm \ref{alg:critic_fixed_utd}
        \For{steps $=1,2,3, \dots, \texttt{UTD}_{actor}$}
            \State Sample $m$ actions from proposal policy $a_1 \dots a_{m} \sim \pi_{\theta'}$
            \State Sort $a_1 \dots a_{m}$ according to $q_\qparams(a_1) \dots q_\qparams(a_{m})$
            \State Set $\mathcal{I} = $ top $\ntop$ actions, $\mathcal{I}' = $ top $2\ntop$ actions
            \State Update policy by gradient ascent using top $\ntop$ actions 
                \Statex \vspace{1em}\centerline{$\displaystyle  \frac{1}{\ntop}\sum_{a\in \mathcal{I}} \nabla_{\pparams} \ln\pi_{\pparams}(a) $}
            \State Update proposal policy by gradient ascent using top $2\ntop$ actions 
                 \Statex \vspace{1em}\centerline{$\displaystyle  \frac{1}{2\ntop}\sum_{a\in \mathcal{I}'} \nabla_{\pparams'} \ln\pi_{\pparams'}(a) $}
        \EndFor
    \EndFor
\end{algorithmic}
\label{alg:ac}
\end{algorithm}

\clearpage
\section{Extended Variability Analysis}
\label{app:extended_results}

\begin{figure*}[p]
	\centering
		\includegraphics[width=0.99\linewidth]{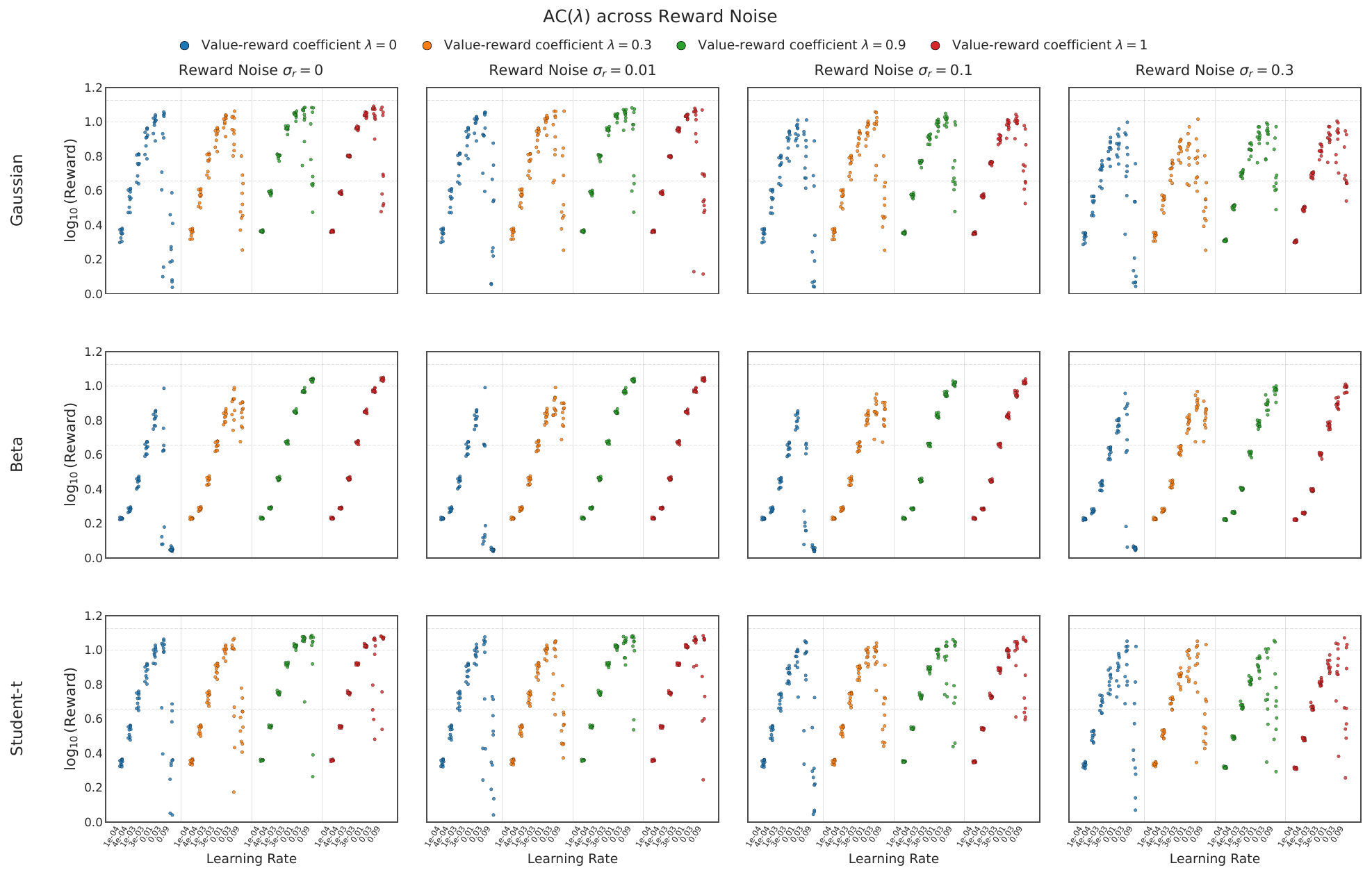}
	\caption
	[Analysis of AC($\lambda$) and changing environmental noise]
	{
		\textbf{Analysis of AC($\lambda$) and changing environmental noise.}
		Each dot represents the average reward of a single training run for a given learning rate.
		The columns represent changing reward noise levels in the environment $\sigma_r$, the rows represent different policy parameterizations, and the colours represent different value-reward mixing coefficients $\lambda$.
		All the other experiments in this paper have used a fixed reward noise of $\sigma_r = 0.01$.
		At $\lambda=1$, the critic is not used in the loss function, whereas at $\lambda=0$, only the critic is used.
		Across all three policies, increasing $\lambda$ shifts the mean performance upwards and reduces the run-to-run variability.
		We observe that for Gaussian and Student's $t$, the best performance is obtained at $\lambda=0.9$, whereas for beta, it is at $\lambda=\{0.9, 1\}$.
		Additionally, unlike other policies, the beta policy's performance does not drop off at the highest learning rate when coupled with a high $\lambda$ value. 
		Increasing environmental noise increases the variability of the Gaussian and Student's $t$ policies, while having no noticeable effect on the beta policy.
	}
	\label{fig:acl_all}
\end{figure*}

\clearpage
\begin{figure*}[p]
	\centering
	\includegraphics[width=0.5\linewidth]{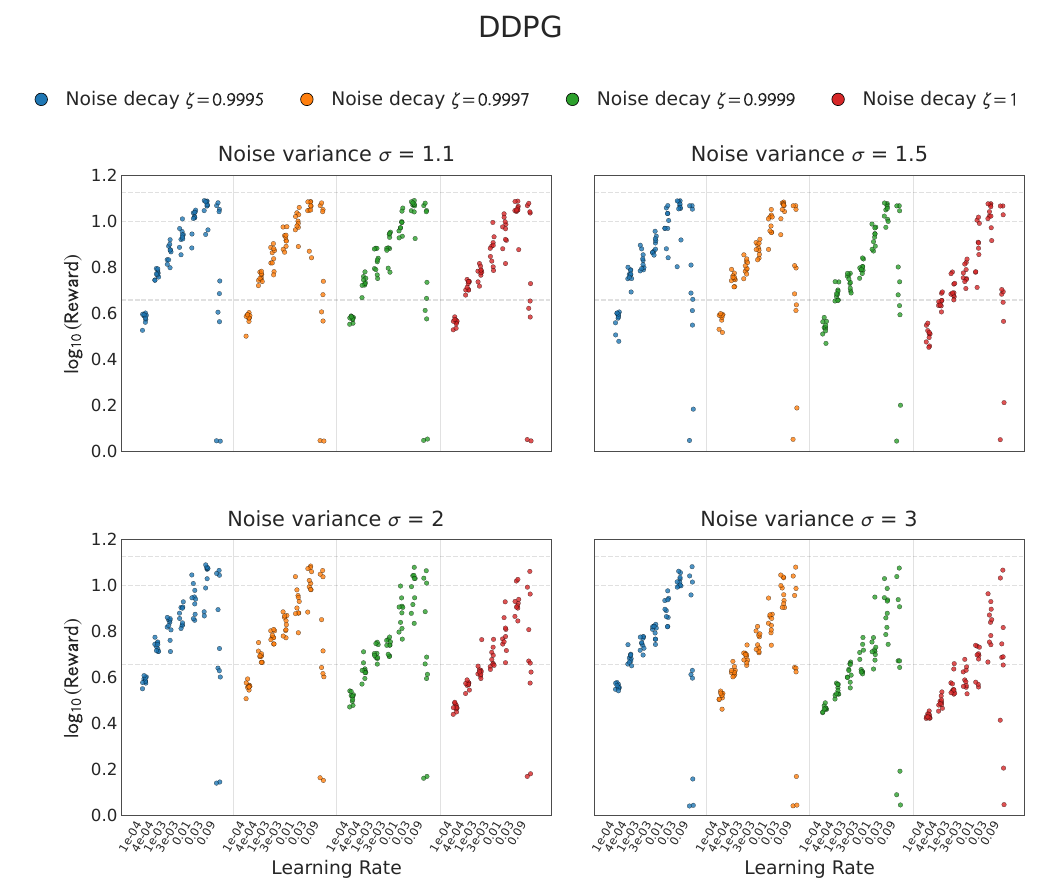}
	\caption
	[Analysis of Deep Deterministic Policy Gradient]
	{
		\textbf{Analysis of Deep Deterministic Policy Gradient.}
		Each dot represents the average reward of a single training run for a given learning rate.
		The 2$\times$2 grid of panels represents different initializations for noise variance $\sigma$ and the colours represent different noise decay rates $\zeta$.
		The noise is decayed at every step as $\sigma = \sigma * \zeta$. 
		We observe that DDPG has lower stability than other algorithms.
		There are very few hyperparameter combinations that lead to a higher mean performance with low variability across runs, and matching the performance of SAC or PPO is difficult to achieve without more extensive tuning of these hyperparameters.
	}
	\label{fig:ddpg_all}
\end{figure*}
\clearpage

\begin{figure*}[p]
	\centering
		\includegraphics[width=0.99\linewidth]{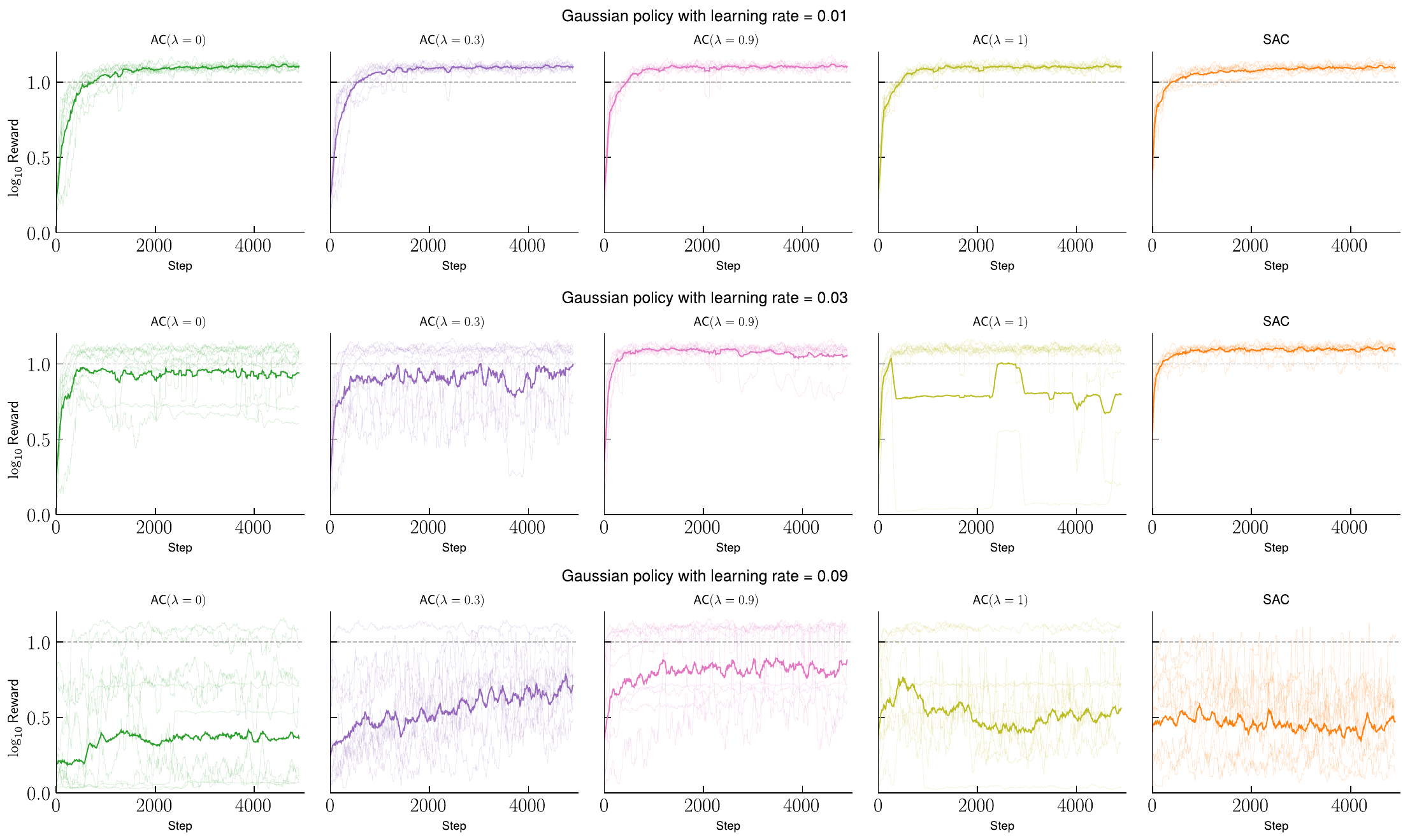}
	\caption
	[Analysis of AC($\lambda$) and Soft Actor-Critic rewards with Gaussian policy]
	{
		\textbf{Analysis of AC($\lambda$) and Soft Actor-Critic rewards with Gaussian policy.}
		Each row represents an experiment with a different learning rate $\eta \in \{0.01, 0.03, 0.09\}$, and each column represents a different algorithm: AC($\lambda$) at four values of $\lambda$ followed by SAC.
		Faint lines represent individual seeds and the bold curve represents the mean.
		At the smallest learning rate $\eta=0.01$, every algorithm reaches a mean reward around 1.0 with low seed spread.
		At $\eta=0.03$, which is the default learning rate used throughout this paper (Table~\ref{tab:common-hparams}), we observe early signs of failure: AC($\lambda=0$) shows individual seeds dropping below 0.5, and AC($\lambda=1$) becomes highly variable across seeds.
		In contrast, AC($\lambda=0.9$) and SAC remain near 1.0 with low spread.
		The behaviour at $\eta=0.09$ (bottom row) shows the same trends in extreme form, with AC($\lambda=0$) collapsing and SAC also becoming highly variable.
		This supports the finding that mixing reward with value estimates improves both performance and variability.
	}
	\label{fig:acl_line}
\end{figure*}

\begin{figure*}[p]
	\centering
	\begin{minipage}{0.99\textwidth}
		\includegraphics[width=0.99\linewidth]{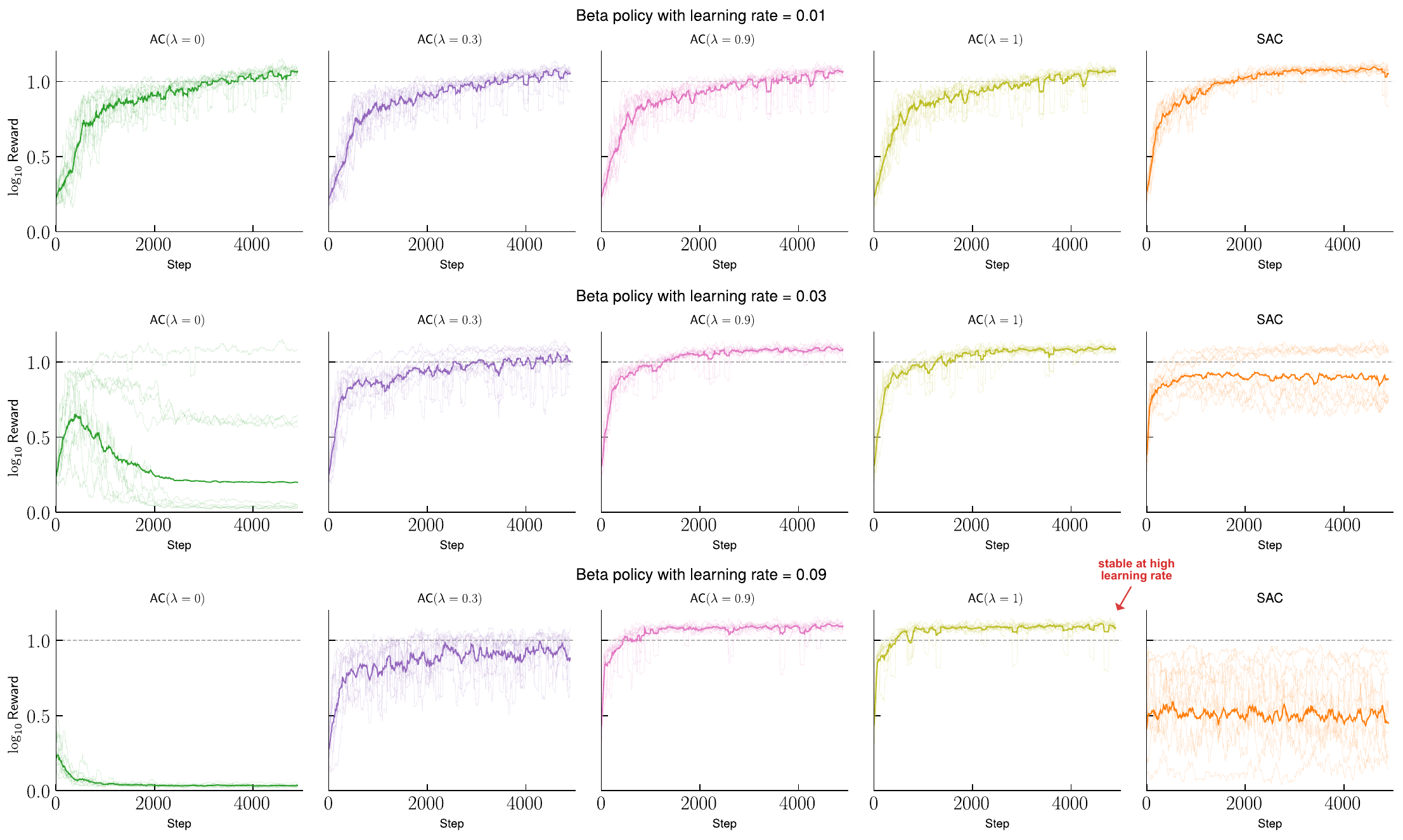}
	\end{minipage}
	\caption
	[Analysis of AC($\lambda$) and Soft Actor-Critic rewards with beta policy]
	{
		\textbf{Analysis of AC($\lambda$) and Soft Actor-Critic rewards with beta policy.}
		Each row represents an experiment with a different learning rate $\eta \in \{0.01, 0.03, 0.09\}$, and each column represents a different algorithm: AC($\lambda$) at four values of $\lambda$ followed by SAC.
		Faint lines represent individual seeds and the bold curve represents the mean.
		We observe a different behaviour as compared to the Gaussian policy in Fig. \ref{fig:acl_line}.
		AC($\lambda=0$) collapses more strongly under the beta policy, with the mean reward dropping to near 0.2 at the default learning rate $\eta=0.03$ (Table~\ref{tab:common-hparams}).
		In contrast, AC($\lambda=0.9$) and AC($\lambda=1$) consistently reach a high mean reward with low run-to-run variability at every learning rate.
		SAC with beta policy can reach the same high reward at a lower learning rate $\eta=0.01$, but becomes highly unstable at higher learning rates, which shows that mixing immediate reward with the critic estimates is useful for lowering run-to-run variability.
		Additionally, Unlike Gaussian policy (Fig. \ref{fig:acl_line}), beta has a very low run-to-run variability at the highest learning rate with $\lambda=\{0.9, 1\}$.
	}
	\label{fig:acl_line_beta}
\end{figure*}

\clearpage
\section{Reproducibility and Disclosures}
\label{appendix:reproducibility}
\subsection*{Generative AI Use Declaration}
\label{appendix:reproducibility}

The authors declare the use of generative AI tools in the research and 
writing process. Following the GAIDeT taxonomy \citep{suchikova2026gaidet}, the 
following tasks were delegated to AI tools under full human supervision:
\begin{itemize}
    \item Code generation
    \item Proofreading and editing
\end{itemize}

\noindent\textbf{Tools used:} Grammarly; Claude Opus 4.7; 
Gemini 3.5.

\noindent\textbf{Scope.}
Grammarly was used to proofread and fix grammatical errors in the manuscript. 
AI assistance with code was limited to adapting existing scripts written by the authors, such as modifying variables, changing data-fetch paths, and providing debugging assistance. All AI-assisted outputs were reviewed and verified by the authors before inclusion.

\subsection*{Experiments}
All experiments use the Backwashing-PID environment with the shared defaults listed in Tables~\ref{tab:env-hypers}, \ref{tab:common-hparams}, and \ref{tab:algo-specific-hparams}.
The codebase uses Python~3.10 with PyTorch as the ML library.
All sources of randomness are controlled by a single global seed per run.
The determinism is enforced by \texttt{torch.use\_deterministic\_algorithms(True)}, \texttt{torch.manual\_seed} and \texttt{np.random.seed}.
Replay buffer sampling uses Python's \texttt{random.sample}, and all environment stochasticity (such as measurement noise $\varepsilon^{(f)} \sim \mathcal{N}(0, 0.003^2)$ and reward noise $\varepsilon^{(r)} \sim \mathcal{N}(0, 0.01^2)$) is drawn via \texttt{np.random.normal}.
Given the same seed, runs are therefore reproducible.
We release the code for all experiments reported in this paper at \url{https://github.com/haseebs/deconstruct-ac}.

%
%
%
\clearpage

\begin{table}[p]
	\centering
	\caption{Environment hyperparameters}
	\label{tab:env-hypers}
	\begin{tabular}{ll}
		\toprule
		Parameter & Value \\
		\midrule
		Cycle length $\mathcal J$ & 55 \\
		Timestep interval $\Delta t$ & 1 \\
		Pump speed bounds $[s_{\min}, s_{\max}]$ & [0, 100] \\
		Measurement noise std $\sigma_f$ & 0.003 \\
		Reward noise std $\sigma_r$ & 0.01 \\
		Deviation norm divisor & 110 \\
		Action space range & $p,i,d \in [0,20]$ \\
		Initial pump speed $s_0$ & 20 \\
        Flow-rate setpoint $f_{ideal}$ & $0.6309013$ \\
		\bottomrule
	\end{tabular}
\end{table}
\clearpage

\begin{table}[p]
	\centering
	\caption{Fixed values for hyperparameters common across the tested algorithms}
	\label{tab:common-hparams}
	\begin{tabular}{l l}
		\toprule
		\textbf{Hyperparameter} & \textbf{Value} \\
		\midrule
		
		Gaussian initialization &
		$\mu$: 0, $\sigma$: 1\\
		
		Beta initialization &
		$\alpha$: 5, $\beta$: 5\\
		
		Student's $t$ initialization &
		$\mu$: 0, $\sigma$: 1, $\nu$: 3\\

		Mini-batch size &
		512 \\
		
		Critic MLP layers &
		2 \\
		
		Critic MLP features &
		64 \\
		
		Adaptive critic UTD error tolerance $\omega$ &
		0.001 \\
		
		Adaptive critic UTD max updates per step $n_\text{steps}$ &
		100 \\
		
		Value baseline decay rate $\eta_v$&
		0.1\\
		
		Actor and critic learning rate $\eta_{actor}$, $\eta_{critic}$&
		0.03\\
		Actor and critic optimizer&
		Adam $\beta:(0.9, 0.999)$\\
		\bottomrule
	\end{tabular}
\end{table}
\clearpage

\begin{table}[p]
	\centering
	\caption{Fixed values for algorithm-specific hyperparameters across all the tested algorithms}
	\label{tab:algo-specific-hparams}
	\begin{tabular}{l l}
		\toprule
		\textbf{Hyperparameter} & \textbf{Default(s)} \\
		\midrule
		\multicolumn{2}{l}{\textbf{PPO}} \\
		Clipping threshold $\epsilon$ & 0.2 \\
		On-policy minibatch size $b$ & 10 \\
		\midrule
		\multicolumn{2}{l}{\textbf{SAC}} \\[2pt]
		Entropy coefficient $\beta$ & 0.001 \\
		Number of actions to sample $m$ & 30 \\
		\midrule
		\multicolumn{2}{l}{\textbf{MPO}} \\[2pt]
		Number of actions to sample $m$ & 30 \\
		\midrule
		\multicolumn{2}{l}{\textbf{DDPG}} \\
		\texttt{Everything is swept} &  \\
		\midrule
		\multicolumn{2}{l}{\textbf{GreedyAC}} \\[2pt]
		Number of actions to sample $m$ & 30 \\
		Number of best actions $n_{top}$ & $0.1*m$ \\
		\midrule
		\multicolumn{2}{l}{\textbf{AC($\lambda$)}} \\[2pt]
		\texttt{Everything is swept} &  \\
		\bottomrule
	\end{tabular}
\end{table}

\clearpage

\begin{table*}[p]
	\caption
	[Overview of experiments and swept hyperparameters and component choices.]
	{Overview of experiments and swept hyperparameters and component choices. Every configuration is run over $10$ random seeds. The learning rate grid $\eta \in \{1.2\!\times\!10^{-4}, 3.7\!\times\!10^{-4}, 1.1\!\times\!10^{-3}, 3.3\!\times\!10^{-3}, 10^{-2}, 3\!\times\!10^{-2}, 9\!\times\!10^{-2}\}$ is swept in every experiment unless noted otherwise. Unless a hyperparameter is listed as swept, it is fixed to the default in Tables~\ref{tab:common-hparams},~\ref{tab:algo-specific-hparams}.
	}
	\centering
	\label{tab:exp-overview}
	\small
	\begin{tabular}{@{}l p{3.0cm} p{7.7cm}@{}}
		\toprule
		\textbf{Experiment} & \textbf{Policy parameterizations} & \textbf{Swept hyperparameters} \\
		\midrule
		\addlinespace[2pt]
		SAC, main sweep
		& Gaussian, squashed Gaussian, beta, Student's $t$, squashed Student's $t$
		& Entropy coefficient $\beta \in \{10^{-5}, 10^{-4}, 10^{-3}, 10^{-2}, 0.1\}$; actor UTD $\texttt{UTD}_{actor} \in \{1, 3, 10, 30\}$; gradient estimator $\in \{\text{pathwise}, \text{likelihood-ratio}\}$ \\
		\addlinespace[3pt]
		SAC, fixed critic UTD
		& Gaussian, beta
		& Critic UTD $\texttt{UTD}_{critic} \in \{1, 10, 100\}$; actor UTD $\texttt{UTD}_{actor} \in \{1, 10\}$ \\
		\addlinespace[3pt]
		SAC, $\eta_{actor}$ vs.\ $\texttt{UTD}_{actor}$ scaling
		& Gaussian, beta
		& Scaling factor $k \in \{1, 3, 10, 30\}$; strategy $\in \{\text{scale }\eta_{actor}, \text{scale }\texttt{UTD}_{actor}\}$; critic LR $\eta_{critic} \in \{3.7\!\times\!10^{-4}, 3.3\!\times\!10^{-3}, 3\!\times\!10^{-2}\}$ (LR axis not swept independently) \\
		\addlinespace[3pt]
		SAC, mirror descent
		& Gaussian
		& Entropy $\beta \in \{10^{-3}, 10^{-2}, 0.1\}$; MD step size $1/\tau \in \{10^{-3}, 10^{-2}, 0.1, 0.5\}$; MD period $\eta_{md} \in \{3, 10, 100, 1000\}$; LR sub-grid of 5 values \\
		\midrule
		\addlinespace[2pt]
		AC$(\lambda)$, main sweep
		& Gaussian, beta, Student's $t$
		& Value--reward mixing coefficient $\lambda \in \{0, 0.3, 0.9, 1\}$ \\
		\addlinespace[3pt]
		AC$(\lambda)$, reward-noise sweep
		& Gaussian, beta, Student's $t$
		& $\lambda \in \{0, 0.3, 0.9, 1\}$; reward noise std $\sigma_r \in \{0, 0.01, 0.1, 0.3\}$ \\
		\addlinespace[3pt]
		AC$(\lambda)$, critic capacity
		& Gaussian, beta, Student's $t$
		& Critic hidden units per layer $\in \{64, 256\}$ (2 hidden layers); $\lambda = 0$ \\
		\midrule
		\addlinespace[2pt]
		PPO, actor-UTD sweep
		& Gaussian, squashed Gaussian, beta, Student's $t$
		& Entropy coefficient $\beta \in \{10^{-3}, 10^{-2}, 0.1\}$; actor UTD $\texttt{UTD}_{actor} \in \{1, 3, 10, 30\}$; on-policy batch size $b = 10$ (fixed) \\
		\addlinespace[3pt]
		PPO, batch-size sweep
		& Gaussian, beta, softmax
		& Entropy coefficient $\beta \in \{10^{-3}, 10^{-2}, 0.1\}$; on-policy batch size $b \in \{1, 3, 10\}$; actor UTD $\texttt{UTD}_{actor} = 1$ (fixed) \\
		\midrule
		\addlinespace[2pt]
		MPO
		& Gaussian, squashed Gaussian, beta, Student's $t$
		& KL coefficient $\tau \in \{0.01, 0.05, 0.1, 0.5\}$; actor UTD $\texttt{UTD}_{actor} \in \{1, 3, 10, 30\}$ \\
		\addlinespace[3pt]
		GreedyAC
		& Gaussian, squashed Gaussian, beta, Student's $t$
		& Actor UTD $\texttt{UTD}_{actor} \in \{1, 3, 10, 30\}$; number of top proposal actions $\eta_{top} \in \{1, 3, 10\}$; proposal samples $m = 30$ (fixed) \\
		\addlinespace[3pt]
		DDPG
		& Deterministic policy with Gaussian exploration noise
		& Initial noise variance $\sigma \in \{1.1, 1.5, 2.0, 3.0\}$; noise decay $\zeta \in \{0.9995, 0.9997, 0.9999, 1.0\}$ \\
		\bottomrule
	\end{tabular}
\end{table*}

\end{document}